\documentclass{article}

\usepackage[final]{neurips_2024}

\usepackage[utf8]{inputenc} %
\usepackage[T1]{fontenc}    %
\usepackage{url}            %
\usepackage{booktabs}       %
\usepackage{amsfonts}       %
\usepackage{nicefrac}       %
\usepackage{microtype}      %
\usepackage{enumitem}
\usepackage{float}
\usepackage{algorithm}
\usepackage[noend]{algpseudocode}

\usepackage[colorlinks=true, allcolors=blue]{hyperref}

\usepackage{amssymb}
\usepackage{amsthm}
\usepackage{xurl}
\usepackage{contour}
\usepackage{amsmath,amsfonts,amssymb}
\usepackage{tablefootnote}
\usepackage[table]{xcolor}
\DeclareMathAlphabet{\mathbbold}{U}{bbold}{m}{n}
\usepackage{multirow}
\usepackage[many]{tcolorbox}    
\usepackage{ragged2e}
\usepackage{xspace}
\usepackage{makecell}
\definecolor{LightCyan}{rgb}{0.7,1,1}

\newcommand*{\Jupdate}[1]{{\textcolor{black}{{#1}}}}

\usepackage[format=plain,
            labelfont=bf]{caption}

\usepackage{hyperref}
\usepackage{cleveref}
\usepackage{wrapfig}
\makeatletter
\def\section{\@startsection {section}{1}{\z@}{-1ex plus -0.2ex minus 0.2ex}{0.1ex plus 0.1ex minus 0.1ex}{\large\bf\raggedright}}
\def\paragraph{\@startsection{paragraph}{4}{\z@}{.1ex plus 0ex minus .1ex}{-1em}{\normalsize\bf}}
\makeatother
\title{Memory Mosaics at scale}

\newcommand{\bfparagraph}[1]{\paragraph{\textbf{#1}}}

\def\eg.{\mbox{e.}\mbox{g.}}
\def\ie.{\mbox{i.}\mbox{e.}}
\def\R{\mathbb{R}}
\def\E{\mathbb{E}}

\def\ttop{{\!\top}}

\def\iid.{\mbox{i.}\mbox{i.}\mbox{d.}}
\def\ood.{\mbox{o.}\mbox{o.}\mbox{d.}}
\definecolor{blue}{HTML}{1a80bb}
\newtcolorbox{boxA}{
    fontupper = \it,
    boxrule = .5pt,
    colframe = black %
}

\newtcolorbox{boxAwhite}{
    boxrule = .5pt,
    colframe = black %
}

\newtcolorbox{boxAdash}{
    fontupper = \it,
    boxrule = .5pt,
    colframe = black %
}

\newtcolorbox{boxB}{
    fontupper = \bf\color{main}, %
    boxrule = 1.5pt,
    colframe = main,
    rounded corners,
    arc = 5pt   %
}

\newtcolorbox{boxC}{
    colback = sub, %
    boxrule = 0pt  %
}

\newtcolorbox{boxD}{
    colback = sub, 
    colframe = main, 
    boxrule = 0pt, 
    toprule = 3pt, %
    bottomrule = 3pt %
}

\newtcolorbox{boxE}{
    enhanced, %
    boxrule = 0pt, %
    borderline = {0.75pt}{0pt}{main}, %
    borderline = {0.75pt}{2pt}{sub} %
}

\newtcolorbox{boxF}{
    fontupper = \it,
    enhanced,
    boxrule = 0.5pt, 
    colframe = white, %
    borderline = {0.5pt}{0pt}{main, dashed} %
}

\newtcolorbox{boxG}{
    enhanced,
    boxrule = 0pt,
    colback = sub,
    borderline west = {1pt}{0pt}{main}, 
    borderline west = {0.75pt}{2pt}{main}, 
    borderline east = {1pt}{0pt}{main}, 
    borderline east = {0.75pt}{2pt}{main}
}

\newtcolorbox{boxH}{
    colback = sub, 
    colframe = main, 
    boxrule = 0pt, 
    leftrule = 6pt %
}

\newtcolorbox{boxI}{
    colback = sub, 
    colframe = main, 
    boxrule = 0pt, 
    toprule = 6pt %
}

\newtcolorbox{boxJ}{
    sharpish corners, %
    colback = sub, 
    colframe = main, 
    boxrule = 0pt, 
    toprule = 4.5pt, %
    enhanced,
    fuzzy shadow = {0pt}{-2pt}{-0.5pt}{0.5pt}{black!35} %
}

\newtcolorbox{boxK}{
    sharpish corners, %
    boxrule = 0pt,
    toprule = 4.5pt, %
    enhanced,
    fuzzy shadow = {0pt}{-2pt}{-0.5pt}{0.5pt}{black!35} %
}

\newtcolorbox{boxL}{
    fontupper = \color{main},
    rounded corners,
    arc = 6pt,
    colback = sub, 
    colframe = main!50, 
    boxrule = 0pt, 
    bottomrule = 4.5pt 
}

\newtcolorbox{boxM}{
    fontupper = \color{white},
    rounded corners,
    arc = 6pt,
    colback = main!80, 
    colframe = main, 
    boxrule = 0pt, 
    bottomrule = 4.5pt,
    enhanced,
    fuzzy shadow = {0pt}{-3pt}{-0.5pt}{0.5pt}{black!35}
}

\author{
        Jianyu Zhang\\
        New York University, New York\\
        FAIR, Meta Inc., New York
        \And
        L\'eon Bottou\\
        FAIR, Meta Inc., New York\\
        New York University, New York
}

\begin{document}

\maketitle

\begin{abstract}
Memory Mosaics \citep{zhang-2025}, networks of associative memories, have demonstrated appealing compositional and in-context learning capabilities on medium-scale networks (\textsc{gpt}-2 scale) and synthetic small datasets. This work shows that these favorable properties remain when we scale memory mosaics to large language model sizes (llama-8\textsc{b} scale) and real-world datasets. 

To this end, we scale memory mosaics to 10\textsc{b} size, we train them on one trillion tokens, we introduce a couple architectural modifications (``\emph{memory mosaics v2}''), we assess their capabilities across three evaluation dimensions: training-knowledge storage, new-knowledge storage, and in-context learning. 

Throughout the evaluation, memory mosaics v2 match transformers on the learning of training knowledge (first dimension) and significantly outperforms transformers on carrying out new tasks at inference time (second and third dimensions). These improvements cannot be easily replicated by simply increasing the training data for transformers. A memory mosaics v2 trained on one trillion tokens still perform better on these tasks than a transformer trained on eight trillion tokens. 

\end{abstract}

\setlength{\tabcolsep}{0.9mm} 

\section{Introduction }

In Machine Learning, compositional capabilities and in-context/out-of-distribution learning capabilities have been continuously pursued but remain challenging. Early attempts to achieve these goals include pursuing disentanglement via various statistical ``independence'' \citep{comon1994independent,roth2022disentanglement}, pursuing out-of-distribution/learning from the perspective of optimization on multiple environments \citep{finn2017model, arjovsky2019invariant,bengio-2019}. In contrast, transformer-based models demonstrate certain compositional capabilities and early in-context learning abilities. However,  we still lack a clear understanding of how current transformers achieve these capabilities, and why earlier models were unable to.

Memory mosaics \citep{zhang-2025}, networks of simple key-value associative memories (without position encoding), offer a comparatively transparent way to understand how composition or disentanglement occur. Trained and evaluated on medium-scale networks and synthetic datasets, memory mosaics reveal promising superior in-context learning abilities. Therefore, we ask \textit{``To peruse a strong and general new task learning capability, how can we scale memory mosaics to large networks and real-world datasets?''}

The contribution of this work: 1) We successfully scale up memory mosaics to {llama}-8\textsc{B} scale, using one trillion real-world training tokens. The resulting network is named as \textbf{Memory Mosaics v2}.\footnote{For clarity, ``Memory Mosaics'' refers to the version in \citet{zhang-2025}, while ``Memory Mosaics v2'' refers to this scaled-up version.} 
\footnote{\Jupdate{\url{https://github.com/facebookresearch/MemoryMosaics}}} 
Compared to memory mosaics, memory mosaics v2 made three architectural modifications, including an adaptive bandwidth of associative memory, a gated time-variant key feature extractor, and a 3-level memory design. 2) We propose three evaluation dimensions to comprehensively assess model ability (from \iid. to \ood. scenarios). 3) Our memory mosaics v2 demonstrate superior new-task learning capabilities (with fewer examples and less priori knowledge from human designers). 

This paper is organized as follows. Section \ref{sec:background} introduces background knowledge on associative memories. Section \ref{sec:mmv2} presents the architecture of memory mosaics v2. Then section \ref{sec:mmv2_training} describes the training process, section \ref{sec:evaluation} evaluates memory mosaics v2 and transformers across three dimensions -- training (persistent) knowledge storage, new knowledge storage, and in-context learning. Section \ref{sec:risk-return-trade-off-mmv2} discusses the failure of replicating memory mosaics v2 by simply increasing the training data ($\times$8 more data) for transformers. \Jupdate{Section \ref{sec:fine-tuning_speed} studies the advantage of memory mosaics v2 in fine-tuning.} Finally, section \ref{sec:discussion} provides discussion and further directions.

\label{sec:memory_mosaics_v2}

\section{Background on Associative Memory}
\label{sec:background}

General speaking, memory mosaics architecture \citep{zhang-2025} replaces attention blocks in transformers \citep{vaswani2017attention} with associative memories. This section provides the background on associative memories, highlights the connection and differences between ``associative memory'' in memory mosaics and ``attention'' in transformers.

\paragraph{Associative Memory}

Associative memories have a long history in both psychology and computer science, referring to relationships between unrelated items. In this work, we follow the definition from \citet{zhang-2025}, according to which an associative memory is a device that can store key-value pairs $\{(k_1,v_1)\dots(k_n,v_n)\}$ and retrieve values given a corresponding key:\footnote{both keys and values shall be vectors in~$\R^d$.}
\begin{equation}
\label{eq:associative_mem}
     k \mapsto  f(k;\: \{(k_1,v_1)\dots(k_n,v_n)\})
\end{equation}

The key-value pairs are stored in a set, and thus can be assumed to be permutation invariant. This exchangeability property suggests that we can view an associative memory as a device that estimates a conditional probability distribution $P(V|K)$ on the basis of the sample $(k_1,v_1)\dots(k_n,v_n)$ of key-value pairs. The retrieval function is then a conditional expectation over this estimated distribution:
\begin{equation}
\label{eq:regression}
     f\big(k;\: \{(k_1,v_1)\dots(k_n,v_n)\}\big) ~=~ \E\/(V\:|\:K=k)\,.
\end{equation}

This conditional expectation can be estimated by kernel regression, e.g. Gaussian kernel regression:\footnote{This Gaussian kernel smoothing not only converges to the true conditional expectation $\E(K|V)$ when $n\rightarrow\infty$ and $\beta=\sqrt{n}$ \citep{nadaraya-1964,watson-1964}, but also makes it easy to compute gradients.}
\begin{equation}
\label{eq:gaussian-smoothing-with-distance}
   f\big(k;\: \{(k_1,v_1)\dots(k_n,v_n)\}\big) ~=~ {\sum_{i=1}^{n} {\footnotesize \frac{e^{-\beta \| k-k_i\| ^2}}{{\sum_{i=1}^{n}} e^{-\beta\| k-k_i \| ^2}}}}~ v_i ~,
\end{equation} 
where $\beta$ controls the bandwidth of Guassian kernel.

\paragraph{Connection between associative memory and attention}

Associative memory in Equation \ref{eq:gaussian-smoothing-with-distance} is closely connected to attention \citep{bahdanau-2015} when all key vectors $k_i$ share the same squared norm. That is, expression~\eqref{eq:gaussian-smoothing-with-distance} becomes: \footnote{For the easy of reading, we reuse the ``attention score'' notion to $\frac{e^{\,\beta\,k^\ttop k_i}}{\sum_{j=1}^{n} e^{\,\beta\,k^\ttop k_j} }$ in associative memories.}
\vspace{-1.5ex}
\begin{equation}
\label{eq:gaussian-smoothing-with-dp}
    f\big(k;\: \{(k_1,v_1)\dots(k_n,v_n)\}\big) ~=~ 
      \sum_{i=1}^{n}
      ~ {\footnotesize \frac{e^{\,\beta\,k^\ttop k_i}}{\sum_{j=1}^{n} e^{\,\beta\,k^\ttop k_j}} } 
      ~ v_i~.
\end{equation}
Moreover, the size of associative memory (i.e., the number of key-value examples) is analogous to the sequence length in attention.

\paragraph{Differences between associative memory and attention} The associative memory viewpoint is conceptually simple and transparent. This simplicity contributes several key differences in associative memory (compared with attention), including: 1) $L_2$ normalized key vectors with an explicit bandwidth parameter $\beta$, 2) a symmetric kernel with the same formula for keys as queries, and 3) the absence of explicit position encoding. These differences further contribute to the superior compositional capabilities and in-context learning capabilities in memory mosaics.

\section{Memory Mosaics v2}
\label{sec:mmv2}

Transformers show an interesting induction head mechanism \citep{olsson2022context}, that is, predict $b$ after sequence $[\dots, a, b, \dots , a]$. This mechanism contributes to their in-context learning ability. 
According to \citet{bietti2023birth}, position encoding and asymmetric query-key extractors (e.g. $q_T=W_qx_T$, $k_T=W_kx_T$) are essential for transformers to achieve this induction head mechanism with at least two layers of attention.

Inspired by studies of induction head mechanism, Memory Mosaics \citep{zhang-2025} construct associative memories using keys to represent the recent past and values to represent the near future (Figure \ref{fig:memory-mosaics-v2-architecture} left):
\vspace{-2ex}
\begin{align}
    k_T = \varphi_{\theta}(x_T, x_{T-1},\dots), \quad v_T = \psi_{\theta}(x_{T+1}, x_T, \dots)
\end{align}
This simple designer allows memory mosaics to get ride of explicit position encoding, use the same key as query, perform induction head with only one layer. The resulting memory mosaics also reveal appealing in-context learning capabilities on small synthetic datasets.

Based on memory mosaics, this section introduces \textbf{Memory Mosaics v2}, aiming at a stronger and more general in-context learning ability on broader real-world tasks (without loss of performance on other common benchmarks). Compared to memory mosaics, memory mosaics v2 incorporates three architecture modifications, including an adaptive bandwidth in associative memory, a gated time-variant key feature extractor, and a 3-level memory design.

\subsection{Adaptive bandwidth in Gaussian kernel smoothing}

Memory mosaics use one fixed bandwidth parameter $\beta$ for different sizes $n$ of associative memory (Equation \ref{eq:associative_mem}). It is well known that bandwidth controls the bias-variance trade-off \citep{hastie2009elements} of kernel regression (memory-based) methods. That is, for a given distribution, the optimal bandwidth depends on the number of examples (key-value pairs in associative memory). Inspired by the asymptotic Mean Integrated Squared Error kernel bandwidth estimation approach where $1/\sqrt{\beta} \propto n^{-1/(p+4)}$ \citep{Garcia-Portugues2024}, memory mosaics v2 schedule $\beta$ in Equation \ref{eq:gaussian-smoothing-with-dp} as:
\begin{equation}
\label{eq:adaptive_bandwidth}
    \beta = \beta_1  n^{\alpha} + \beta_0\,,
\end{equation} 
where $\beta_0\geq0,\,\beta_1>0,\,1>\alpha>0$ are learnable parameters (Check Appendix Table \ref{tab:param_init} for reparameterization and initialization details). I.e., the more key-value pairs (examples), the smaller bandwidth $1/\sqrt{\beta}$.

\subsection{Gated time-variant key feature extractor} 

Memory mosaics employs a simple time-invariant leaky averaging to extract key features:
\vspace{-0.5ex}
\begin{equation}
\label{eq:mm_key_extractor}
 \begin{aligned}
 k_T &= \mathrm{Norm}\big(\textcolor{purple}{\Bar{k}_T}\big) &\text{with}  \quad {\textcolor{purple}{\Bar{k}_T} = \textcolor{blue}{\tilde{k}_T} + \lambda\, \textcolor{purple}{\Bar{k}_{T-1}} \quad\quad \textcolor{blue}{\tilde{k}_T} = W_\varphi\,x_T  }
 \end{aligned}
\end{equation}
The averaging weights in Equation \ref{eq:mm_key_extractor} are fixed and independent of the semantic input $x$. As a result, semantically similar cases, such as ``tom-and-jerry'' and ``tom- - -and- - -jerry'', may receive different key features. Inspired by recurrent-style networks \citep{peng2023rwkv,gu2023mamba, beck2025xlstm}, memory mosaics v2 utilize the following gated time-variant key feature extractor: \footnote{It worth noting that this work is neither a linearization of attention nor attention efficiency. The recurrent feature extractor in Eq. \ref{eq:mm_key_extractor} is used to create keys, while associative memory in Eq. \ref{eq:associative_mem} still stores all key-value pairs. }
\begin{equation}
\label{eq:betterfeatures}
 \begin{aligned}
 k_T &= \mathrm{Norm}\big(\textcolor{purple}{\Bar{k}_T}\big) \quad \text{with}  \quad
 \begin{cases}
  {\, \textcolor{purple}{\Bar{k}_T} = g_T\textcolor{blue}{\tilde{k}_T} + \lambda_{T}\textcolor{purple}{\Bar{k}_{T-1}} \quad\quad \textcolor{blue}{\tilde{k}_T} = W_\varphi\,x_T  }\\
  \,   g_t = e^{W_g x_T} \in \mathbb{R}\,,\, \lambda_T = e^{- |W_{\lambda}  x_T|}\in \mathbb{R}
 \end{cases},
 \end{aligned}
\end{equation}
where $W_{\varphi}, W_{g}, W_{\lambda}$ are learnable parameters, the averaging weights $\lambda_T \in \mathbb{R}$ and the exponential gate $g_T\in \mathbb{R}$ semantically depend on input $x_T$. See Appendix Figure \ref{fig:key-feature-extractor-gated} for graphical illustrations. 

For key feature extractor, memory mosaics v2 reuses the same convolutional key extractor as in memory mosaics: 
\vspace{-0.6ex}
\begin{equation}
\label{eq:betterfeatures}
 \begin{aligned}
   v_T &= \alpha_{\psi}\mathrm{Norm}\big(\textcolor{purple}{\Bar{v}_T}\big) \quad\, \text{with}  \quad   {\textcolor{purple}{\Bar{v}_T} = \gamma\, \textcolor{blue}{\tilde{v}_T} + (1-\gamma )\,\textcolor{blue}{\tilde{v}_{T+1}} \quad~ \textcolor{blue}{\tilde{v}_T} = W_\psi\,x_T\,,
 }
 \end{aligned}
\end{equation}
where $\gamma, \alpha_{\psi} \in \mathbb{R}$ and $W_{\psi}$ are learnable parameters.

\subsection{3-level memory}
\label{sec:3-level_memory}
\begin{figure}[ht!]
    \centering
    \vspace{-2ex}
\includegraphics[width=0.37\textwidth]{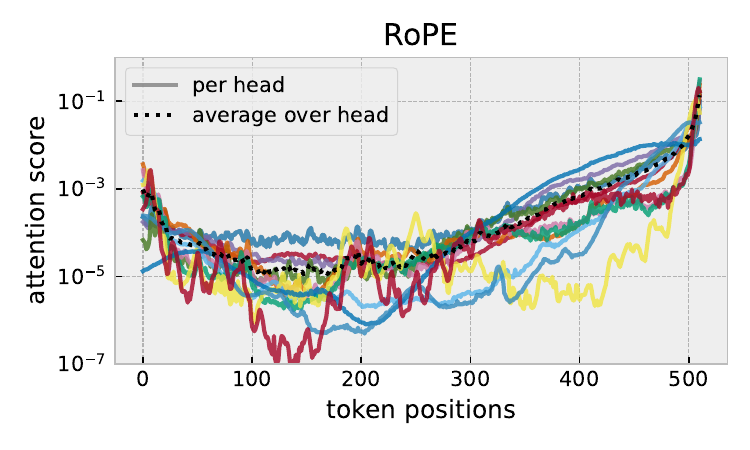} \quad \quad \quad
    \includegraphics[width=0.37\textwidth]{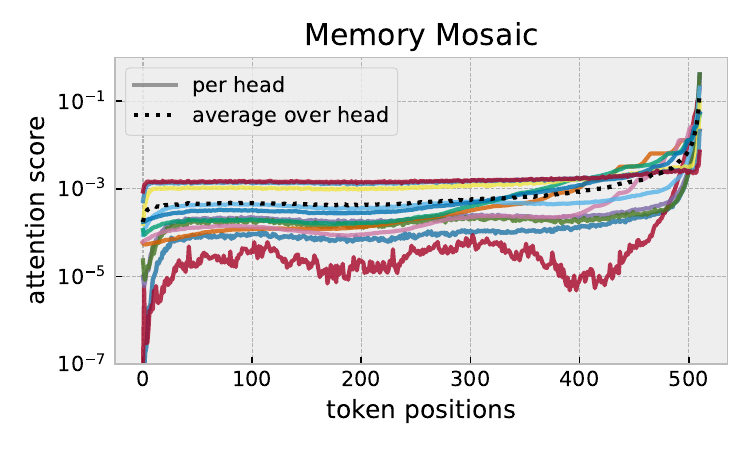}
     \vspace{-1ex}
    \caption{\underline{Average} attention scores of the last token attending previous tokens. \textbf{Left}: Transformer with RoPE position encoding. \textbf{Right}: Memory Mosaics \citep{zhang-2025}. The (averaged) attention scores in transformer heavily depends on token positions (curly curves), while the attention scores in memory mosaics at far tokens (e.g. position 0 to 450) are almost invariant to positions (flat curves). }
    \label{fig:transformer_mm_attn}
     \vspace{-1.5ex}
\end{figure}

\begin{figure}[ht!]
    \centering
     \vspace{-1ex}
    \includegraphics[height=0.34\linewidth]{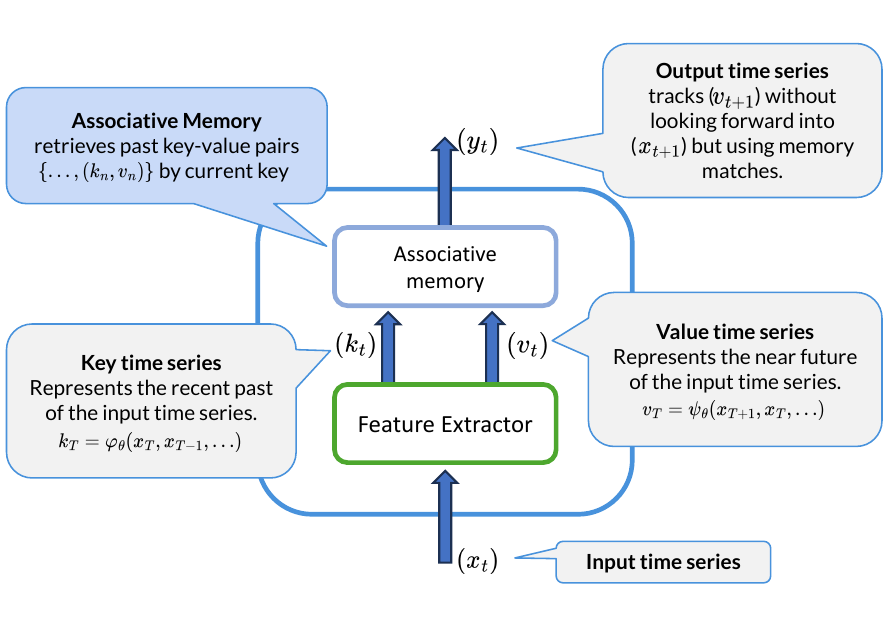} 
    \includegraphics[height=0.34\linewidth]{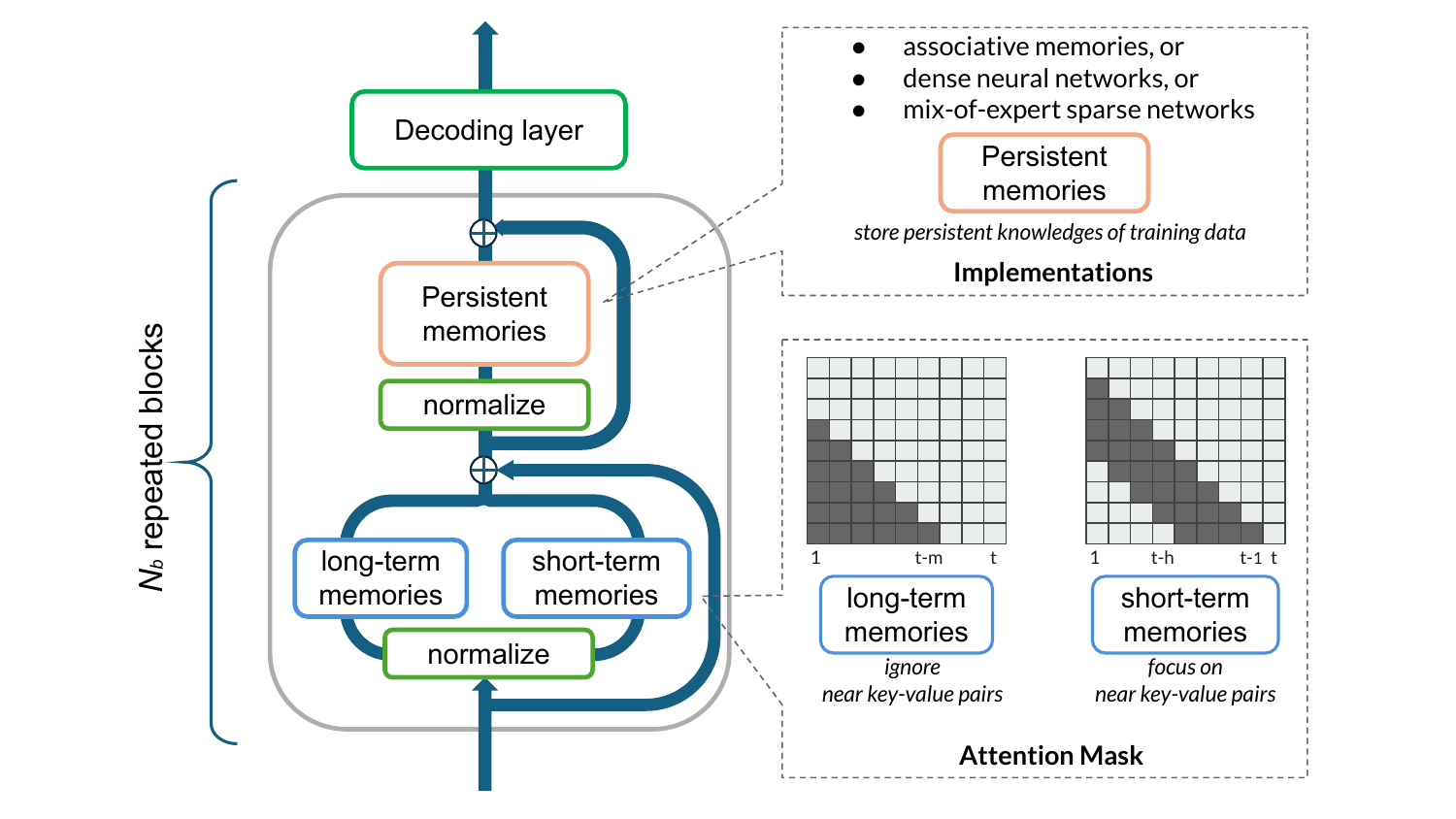}
     \vspace{-1ex}
    \caption{\textbf{Left:} Memory Unit. \textbf{Right:} Memory Mosaics v2 architecture. } %
    \label{fig:memory-mosaics-v2-architecture}
     \vspace{-1ex}
\end{figure}

Transformer architecture \citep{vaswani2017attention} consists of attention blocks and feedforward neural network blocks. The former handles local contextual information from an input sequence, while the latter stores global persistent information shared by different training sequences. Memory mosaics \citep{zhang-2025} simplify the attention and the feedforward network in transformer as contextual associative memory and persistent memory, respectively. This simplification reduces the dependence between the ``attention score'' and the token position, as shown in Figure \ref{fig:transformer_mm_attn}. 
Compared with transformers (Figure \ref{fig:transformer_mm_attn} left), the attention scores in memory mosaics (Figure \ref{fig:transformer_mm_attn} right) exhibit a structured pattern. That is, attention scores on near-tokens (positions) heavily depend on positions, while attention scores on far-tokens are almost invariant to token positions. Inspired by this experimental discovery, memory mosaics v2 replace each contextual associative memory in memory mosaics with two associative memories, \textit{short-term memory} and \textit{long-term memory}, using distinct parameters (as in Figure \ref{fig:memory-mosaics-v2-architecture}). 
\Jupdate{\paragraph{Short-term memory} The short-term memory at position $t$ only stores key-value pairs of near-tokens, ranging from $t-h+1$ to $t-1$, implementing Eq. (\ref{eq:gaussian-smoothing-with-dp}) as:}
\vspace{-1ex}
\begin{equation}
    f\big({k};\: \{({k}_{t-h+1}, {v}_{t-h+1})\dots ({k}_{t-1}, {v}_{t-1})\}\big)~=~ \sum_{i=t-h+1}^{t-1}
      ~ {\footnotesize \frac{e^{\,{\beta}\,{k}^\ttop {k}_i}}{\sum_{j=t-h+1}^{t-1} e^{\,{\beta}\,{k}^\ttop {k}_j}} } 
      ~ {v}_i~.
\end{equation}
\paragraph{Long-term memory} In contrast, the long-term memory skips near tokens and only stores key-value pairs before position $t-m$, implementing Eq. (\ref{eq:gaussian-smoothing-with-dp}) as $f(\textcolor{blue}{k};\: \{(\textcolor{blue}{k}_{1}, \textcolor{blue}{v}_{1})\dots (\textcolor{blue}{k}_{t-m}, \textcolor{blue}{v}_{t-m})\})$.\footnote{Note that $k$, $v$, and $\beta$ in long-term and short-term memory are constructed with distinct parameters.} By setting $m < h$, memory mosaics v2 create an overlap between long-term and short-term memory, resulting in a soft boundary between these two memories. \Jupdate{Eventually, the outputs of many long-term memories and short-term memories are concatenated together, following by a linear projection $W_o$.}

\paragraph{Persistent Memory} Memory mosaics v2 implements \emph{persistent memory} using dense two-layer neural networks with SwiGLU activation \citep{shazeer2020glu} due to computational efficiency concerns.\footnote{A two-layers feed-forward network and a key-value associative memory are interchangeable as shown in \citet{sukhbaatar-2019}.} 

\section{Training}
\label{sec:mmv2_training}

We train two memory mosaics v2 of difference sizes (small/large). Memory mosaics v2 small (llama-1.5\textsc{b} scale) contains 24 layers, 2048 hidden dimensions, and 16 heads, trained on 200 billion tokens of a diverse datamix. Memory mosaics v2 large (llama-8\textsc{b} scale) increases the number of layers to 32, hidden dimensions to 4096, and the number of heads to 32, trained on 1 trillion tokens of the same datamix. Both models are trained on 4,096 context length, followed by a fine-tuning process on 32,768 context length. \Jupdate{Other training details are provided in Appendix \ref{sec:training_details}.}
\Jupdate{\bfparagraph{Stochastic long-term memory size}
During training, memory mosaics v2 samples the long-term memory delay step $m$ from $[64, 256]$, sets the short-term memory window size $h=256$. At inference, $m$ is set to 64. This stochastic long-term memory training setup encourages the allocation of position-invariant signals to long-term memory and position-dependent signals to short-term memory (as shown in Figure \ref{fig:transformer_mm_attn}). The experimental results in Appendix \ref{apx:new-knowledge} Table \ref{tab:stochastic_long_mem} show that this training setup enhances context-length extrapolation ability by more than 15\%.}

\bfparagraph{Baseline}
We train two baseline transformers (small/large) with the same configurations as their memory mosaics v2 counterparts. Unless otherwise specified, in this work, transformer models use llama architecture \citep{grattafiori2024llama} with multi-head attention.

\section{Three evaluation dimensions}
\label{sec:evaluation}

The evaluation design provides a means to assess the specific properties of a system. Memory mosaics v2 aims at the ability to learn new tasks with fewer examples and less task-specific priori knowledge \citep{zhang2025ai}. Thus, to fully assess this capability, this section adopts three evaluation dimensions.
\begin{itemize}[leftmargin=.2in]
     \setlength\itemsep{-0.1em}
    \item \textbf{Persistent-knowledge storage and retrieval}, the ability of persistent-memory to store and retrieve knowledge of training dataset. This capability prepares knowledge that could be reused in other tasks during inference. We use common language benchmarks to access this aspect. 
    \item \textbf{New-knowledge storage and retrieval}, the ability to store and retrieve new information of test dataset. It is a prerequisite for ``learning'' new tasks via memory-based methods. 
    \Jupdate{We employ ``multi-unrelated-documents storing and question-answering'' tasks to evaluate this aspect.}
    
    \item \textbf{In-context Learning}, directly evaluates the ability to learn new tasks with fewer examples and less task-specific priori knowledge. We use multiclass classification to assess this aspect.
\end{itemize}

\subsection{Persistent-knowledge storage and retrieval }
\label{sec:eval_training_data_stsoring_retrieval}

Table \ref{tab:tf_mm_19_common_tasks} evaluates both memory mosaics v2 and baseline transformers on 19 commonly used language benchmarks, showing that they perform closely on these benchmarks.%
This is expected since both models share the same persistent memory architecture. 

\begin{table}[ht!]
 \vspace{-1.5ex}
    \centering
                \caption{Memory mosaics v2 and transformers performance on 19 common language benchmarks. }
    \vspace{0.5ex}
    \label{tab:tf_mm_19_common_tasks}
    \resizebox{\textwidth}{!}{
    \begin{tabular}{cc|ccc ccc ccc c ccc ccc ccc |c}
    \toprule
model        & \makecell{context\\length}         &   obqa  &  \makecell{arc\\easy}  &   \makecell{wino-\\grande}   &  \makecell{arc\\challenge} &  piqa  &  boolq  &  \makecell{hell-\\aswag}  &  nq  &  siqa  &  tqa  &  gsm8k  &  \makecell{mmlu\\alt}   &  \makecell{human \\eval+}  &  squad  &  bbh  &  math  &  mbpp  &  \makecell{race\\middle}  &  \makecell{race\\high}  &   avg \\
\midrule
transformer small & 32k     &  35.2  &  61.0  &  60.1  &  31.4  &  73.6  &  63.0  &  59.3  &  11.7  &  44.5  &  26.7  &  3.0  &  35.2  &  32.4  &  54.7  &  26.0  &  1.2  &  9.2  &  52.2  &  37.4  & \cellcolor{blue!25} 37.8 \\
memory mosaics v2 small & 32k  &  35.0  &  60.0  &  58.4  &  32.9  &  73.3  &  62.7  &  58.0  &  11.8  &  46.6  &  29.3  &  3.1  &  34.7  &  30.8  &  59.3  &  27.3  &  1.1 &  9.4  &  49.2    &  38.4  &  \cellcolor{blue!25}  38.0 \\
\midrule
transformer large & 32k     &  45.8  &  77.3  &  72.3  &  52.6  &  80.8  &  72.6  &  79.2  &  31.9  &  49.3  &  61.5  &  32.4  &  49.0  &  38.3  &  76.3  &  45.6  &  8.7  &  9.8  &  62.6  &  45.6  & \cellcolor{blue!25} 52.2 \\
memory mosaics v2 large & 32k  &  45.4  &  78.0  &  71.2  &  51.8  &  80.4  &  73.1  &  78.6  &  30.9  &  48.6  &  62.0  &  27.4  &  48.2  &  43.0  &  78.2  &  47.8  &  8.8  &  9.6  &  61.6  &  46.5  & \cellcolor{blue!25}  52.2 \\
\bottomrule
    \end{tabular}
    }
 \vspace{-1ex}
\end{table}

How do we know whether these benchmarks access persistent-knowledge ability rather than new-knowledge ability? To answer this question, we re-evaluate these benchmarks on memory mosaics v2 but with \emph{long-term memory} being removed after training.  The underlying reason is that if a task solely relies on the information stored in persistent memory and retrieved by {short-term memory}, removing long-term memory should not significantly affect performance.

Table \ref{tab:13_normal_benchmarks_with_without_longterm_mem} shows that removing {long-term memory} after training does not degrade the performance of 13 common benchmarks. 
This suggests that these 13 tasks are almost exclusively based on information stored in persistent memory and retrieved by short-term memory. In contrast, Appendix Table \ref{tab:6_normal_benchmarks_with_without_longterm_mem} indicates that the other 6 benchmarks perform poorly when long-term memory is removed.

Based on these findings, we use the 13 tasks to evaluate persistent knowledge storage and retrieval capability. The results (Table \ref{tab:tf_mm_19_common_tasks}) show that memory mosaics v2 and transformers perform similarly in this evaluation dimension, suggesting that both models are capable of effectively storing and retrieving persistent knowledge.

\begin{table}[ht!]
 \vspace{-1.5ex}
    \centering
       \caption{Memory mosaics v2 performance on 13 common language benchmarks. Removing the ``long-term memory'' after training barely hurt the performance (56.6\% vs 56.8\%). Flops/token is estimated at context length 256 via tha approach of \citet{casson2023transformerflops}.  }  
        \vspace{0.5ex}
    \label{tab:13_normal_benchmarks_with_without_longterm_mem}
    \resizebox{0.92\textwidth}{!}{
    \begin{tabular}{c|cc|ccc ccc ccc c ccc | c  }
    \toprule
  &  params & flops/token  &   obqa  &  \makecell{arc\\easy}  &  \makecell{wino-\\grande}  &  \makecell{arc\\challenge}  &   piqa  &  boolq  &  \makecell{hell-\\aswag}  &  nq  &  siqa  & tqa  & gsm8k  &  \makecell{mmlu\\alt}  &  \makecell{human \\eval+} &  avg \\
         \midrule
         Transformer large  & 8.8B & 16.7B  & 45.8  &  77.3  &  72.3  &  52.6  &  80.8  &  72.6  &  79.2  &  31.9 & 49.3  &  61.5  &  32.4  &  49.0  &  38.3  &  \cellcolor{blue!25} 57.1 \\
         \midrule
         memory mosaics v2 large  & 9.9B &  18.9B  &  45.4  &  78.0  &  71.2  &  51.8   &  80.4  &  73.1  &  78.6  &  30.9  &  48.6 &  62.0  &  27.4  &  48.2  &  43.0  & \cellcolor{blue!25} 56.8 \\
         \makecell{memory mosaics v2 large \\without long-term memory} & 8.3B &15.6B  & 45.4  &  77.9  &  71.2  &  51.8  &  80.4  &  73.1  &  78.6  &  30.8 &  48.6  &  62.1  &  26.7  &  46.8  &  42.2  &  \cellcolor{blue!25} 56.6 \\
         \bottomrule
    \end{tabular}
    }
 \vspace{-1ex}
\end{table}

\paragraph{Computation and \# parameters concerns}

Table \ref{tab:13_normal_benchmarks_with_without_longterm_mem} summarizes the size of parameters and computation required for transformers and memory mosaics v2. \Jupdate{Interestingly, removing long-term memory from memory mosaics v2 after training achieves a comparable transformer performance on the 13 persistent-knowledge benchmarks, while using fewer parameters and computations.}

\subsection{New-knowledge storage and retrieval }
\label{sec:long-term-mem_utilization}

The new-knowledge storage and retrieval ability is a prerequisite for learning new tasks via memory-based methods (e.g., Gaussian kernel regression), because the data of new tasks must be adequately ``stored'' before learning (Note that memory-based methods are lazy methods). To illustrate this point, consider a poor goldfish with 7-second memory -- how can it possibly learn a 90-minute movie? Similarly, a model with limited new-knowledge storage ability will struggle to learn information that exceeds its storage (memory) capacity. 

\paragraph{Task description} To assess this ability, we employ two \Jupdate{``multi-unrelated-documents question-answering''} tasks from the \textsc{ruler} benchmark \citep{hsieh2024ruler}. These tasks involve multiple concatenated realistic articles followed by a question related to one of these articles, requiring the model to find the correct answer based on the correct article.\footnote{Similarly to the process used in section \ref{sec:eval_training_data_stsoring_retrieval} for verifying persistent-knowledge storage and retrieval tasks, appendix Table \ref{tab:tf_mm_32k_ruler_qa_tasks_long-term-mem} compares memory mosaics v2 with and without long-term memory on these question-answering tasks, confirming the necessity of ``long-term memory'' for these tasks.} A prompt example is:
    
\begin{boxAwhite}
    \vspace{-1ex}
    {\footnotesize Answer the question based on the given documents. The following are given documents. Document 1: [...] Document2: [...] [...] Document 20: [...] Question: What religion were the Normans? Answer:}
    \vspace{-1ex}
\end{boxAwhite}
\Jupdate{These tasks are notably more challenging than typical 'needle-in-a-haystack' benchmarks \citep{kamradtnish}, owing to their high information entropy. The typical 'needle-in-a-haystack' task is too easy, resulting in many models achieving near-perfect performance. See Table \ref{tab:ruler_s_niah} in Appendix for details. }

\paragraph{Main results} Table \ref{tab:tf_mm_4k_ruler_qa_tasks} compares memory mosaics v2 and transformers, pretrained on a 4k context length, on these question-answer tasks. Memory mosaics v2 outperforms transformers on 4k task-length by 1.4\%$\sim$5.6\%. Similarly, Table \ref{tab:tf_mm_32k_ruler_qa_tasks} presents the same comparison, but with both models fine-tuned at a 32k context length. As task lengths increase to 32k, the \Jupdate{``multi-unrelated-documents question-answering''} tasks become more challenging. At this increased difficulty level, memory mosaics v2 significantly outperforms transformers by 12.3\% to 14.8\%.

\begin{table}[ht!]
    \centering
     \vspace{-1.5ex}
        \caption{Comparison of memory mosaics v2 and transformer, trained on 4k context length, on \textsc{ruler} question-answer tasks. Memory mosaics v2 not only outperforms transformer on 4k task-length, but also successfully extrapolate the context length $\times4\sim\times8$ times without any fine-tuning.}
    \label{tab:tf_mm_4k_ruler_qa_tasks}
     \vspace{0.5ex}
    \resizebox{0.71\textwidth}{!}{
    \setlength{\tabcolsep}{2mm} 
    \begin{tabular}{cc|c| c c c cc  }
    \toprule
         model          &   \makecell{context length}      &   \makecell{task-length 4k}   &    \makecell{ 8k}   &    \makecell{16k}  &    \makecell{ 32k} && \\
         \midrule
        transformer small       &4k &  39.4  &  $\times$  & $\times$  & $\times$   &&\\
        memory mosaics v2 small    &4k &  45.0  &  35.0  &  34.1  &  31.7 && \\
        \midrule
        transformer large     &4k   &  57.7  &  $\times$  &  $\times$  &  $\times$ &&\\ 
        memory mosaics v2 large  &4k   &  59.3  &  48.8  &  46.4  &  26.5 && \\
\bottomrule
    \end{tabular}
    }
     \vspace{-1ex}
\end{table}
\begin{table}[ht!]
    \centering
     \vspace{-1.5ex}
        \caption{Comparison of memory mosaics v2 and transformer, trained on 4k and fine-tuned on 32k context length, on \textsc{ruler} question-answer tasks. Memory mosaics v2 outperforms transformer by 12.3\%$\sim$14.8\%. }
    \label{tab:tf_mm_32k_ruler_qa_tasks}
     \vspace{0.5ex}
    \resizebox{0.7665\textwidth}{!}{
    \setlength{\tabcolsep}{2mm} 
    \begin{tabular}{cc|c c c | c | c cc }
    \toprule
         model          &   \makecell{context length}      &   \makecell{4k}   &    \makecell{8k}   &    \makecell{16k}  &    \makecell{task-length 32k}  & \makecell{64k}  &&\\
         \midrule
        transformer small    &32k  &  37.0  &  29.3  &  29.0  &  22.1 & $\times$ \\
        memory mosaics v2 small &32k  &  44.3  &  39.3  &  39.4  &  36.9 &  25.3\\
        \midrule
        transformer large    &32k &  51.2  &  48.8  &  44.7  &  41.1 & $\times$ \\
        memory mosaics v2 large &32k &  58.9  &  55.5  &  54.9  &  53.4 &  46.4\\
        \bottomrule
    \end{tabular}
    }
     \vspace{-0.5ex}
\end{table}
\bfparagraph{The failures of many potential baselines}
\label{parag:failues_of_many_baselines}

Many memory compression algorithms, such as \textsc{rnn}s, x\textsc{lstm} \citep{beck2025xlstm}, rwkv \citep{peng2023rwkv}, and state-space models \citep{gu2023mamba}, fail on this task by construction because they cannot store all articles before reading the question. Similarly, local-window memory approaches, such as Alibi position encoding \cite{press2021train} and sliding-window attention \cite{beltagy2020longformer}, also struggle for the same reason.\footnote{One might argue to play around this shortage by reading the question before the multiple articles. However, this process involves task-specific priori knowledge from human designers. In the end, instead of proving the machine is intelligent, it often proves that human designers are intelligent. Please recall that \emph{a child does not prepare all questions before going to school}.} \Jupdate{This incompetent of memory compression algorithms has also been experimentally demonstrated by \citet{hsieh2024ruler} and \citet{li2024long}. Also, see Appendix \ref{apx:new-knowledge} for these experimental evidences.}
\bfparagraph{Extrapolating context length (without fine-tuning)} 
Context length extrapolation (without fine-tuning) not only is computationally appealing, but also reveals the model's consistency in handling context. Unfortunately, transformers (with \textsc{rope} position encoding) struggle to extrapolate context length, as shown in Table \ref{tab:tf_mm_4k_ruler_qa_tasks}.\footnote{The comparison ignores many memory compression and local window approaches \citep{press2021train,beltagy2020longformer}, because they fail on this evaluation by construction.} In contrast, memory mosaics v2, trained on 4k context length, not only outperform transformers on 4k length, but also perform well after extrapolating context length $\times4\sim\times8$ times without any fine-tuning or adaptation.

\subsection{In-context learning}
\label{sec:inference-data-learning}
Having demonstrated the new-knowledge storage and retrieval ability of memory mosaics v2, this section takes a step further to evaluate its capacity to learn new tasks or distributions at inference time. This ability is also commonly referred to as in-context learning. 

\bfparagraph{Tasks description} To assess the in-context learning ability, we employ classic multiclass classification problems,\footnote{We choose classic classification problems over other fancy benchmarks for two reasons. Firstly, the mechanisms underlying classification are well-studied, allowing us to confidently attribute good or poor performance to the system's properties. Secondly, classification tasks can be designed to be arbitrarily different from the training set by changing the classification boundary, making it easier to measure the ability to learn new distributions. In contrast, as of this writing, many fancy benchmarks may not offer the same level of control and fine-grained analysis.} adopted from \citet{li2024long}. The classification tasks include:
\begin{itemize}[leftmargin=.2in]
\vspace{-1ex}
     \setlength\itemsep{-0.1em}
    \item \textbf{Banking77} \citep{casanueva2020efficient} is a banking-intent classification task with 77 target categories. Each example has an average length of 24 tokens. 
    \item \textbf{Tacred} \citep{zhang2017tacred} is a relation classification task of two objects in a sentence, extracted from newswire or webtext, with 41 target categories. Each example has an average length of 77 tokens. 
     \item \textbf{Goemotion} \citep{demszky2020goemotions} is an emotion classification task of Reddit comments with 28 target categories. Each example has an average length of 26 tokens. 
\end{itemize}

To solely evaluate the ability to learn new tasks (reduce the influence of training knowledge), we create an anonymous version with anonymous target labels (e.g. ``class 1'', ``class 2'') for each classification task. The original classification setup with semantic labels (e.g. ``happy'', ``angry'') is referred to as semantic version.

In this section, we adopt a few-shot learning setup where each ``shot'' consists of one $(x,y)$ example from each possible target label category. By collecting multiple shots, we create an $n$-shot classification task. To encode these $(x,y)$ examples for memory mosaics v2 and transformers, we serialize the $(x,y)$ pairs into a sequence followed by a test query $x_{test}$.\footnote{Transformers are known to be sensitive to the prompt strategies \citep{gupta_changing_2024,mirzadeh_gsm-symbolic_2024}, such as the delimiter before $x$ and $y$, shuffling/not-shuffling the $(x,y)$ examples within each shot. To reduce the influence of prompt strategies, we evaluate each classification task with different delimiters (``[space]'' and ``$\backslash n$''), shuffled/non-shuffled $(x,y)$ examples. Then choose the best prompt strategy for each $n$-shot classification task. Check appendix \ref{apx:prompt_examples_icl_class} for prompt examples.} A prompt example is:
\begin{boxAwhite}
    \vspace{-1ex}
    {\footnotesize Given a customer service query, please predict the intent of the query. [...]  The examples are as follows: query: $x_{shot1}$, instant: $y_{shot1}$, [...], query: $x_{shot2}$, instant: $y_{shot2}$, [...], query: $x_{test}$, instant:}
    \vspace{-1ex}
\end{boxAwhite}

\bfparagraph{Main Results} 

\begin{figure}[ht!]
    \centering
        \vspace{-1ex}
    \includegraphics[width=0.92\linewidth]{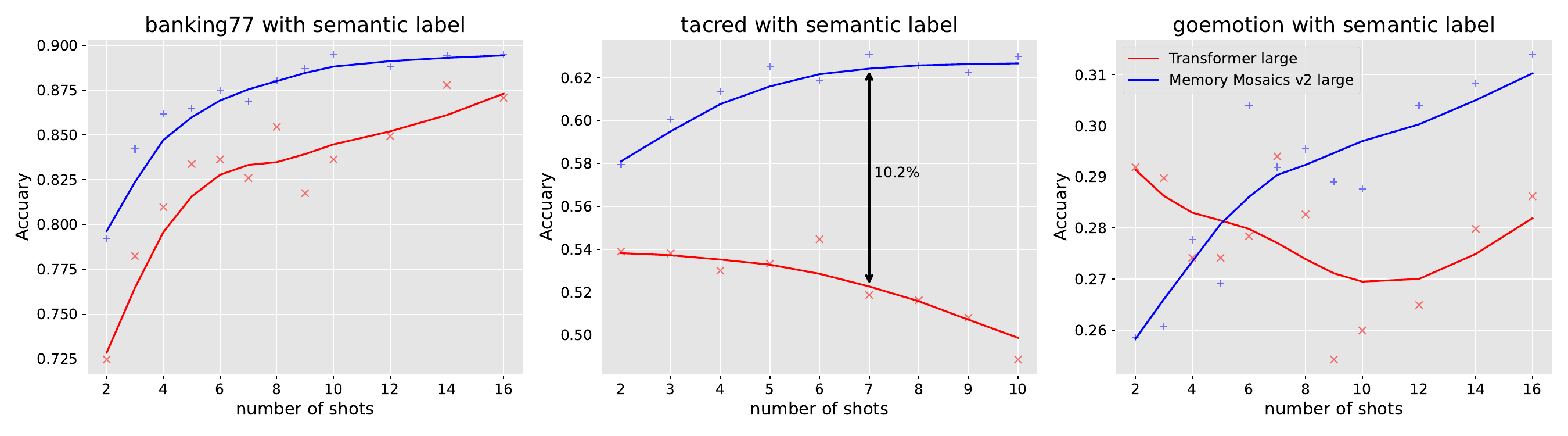}
        \vspace{-1.5ex}
    \caption{Semantic label in-context learning comparison between memory mosaics v2 and transformer. Memory mosaics v2 significantly outperform transformers on in-context learning with a large margin (more than 10\%). Meanwhile, memory mosaics v2 benefits from more demonstration shots (x-axis), unlike transformers.}
    \label{fig:inference-time-classification_semantic_labels}
        \vspace{-1ex}
\end{figure}

Figure \ref{fig:inference-time-classification_semantic_labels} compares the performance of memory mosaics v2 and transformers in three classification tasks with semantic target labels. The horizontal axis represents the number of shots, while the vertical axis represents the classification accuracy on $x_{test}$. We can observe two phenomena: \textbf{1)} memory mosaics v2 consistently improve classification performance as it sees more demonstration shots (\textcolor{blue}{blue curves}). In contrast, transformers struggle to maintain their performance and exhibit counterintuitively degraded performance as more demonstrations are provided (\textcolor{red}{red curves}). \textbf{2)} Memory mosaics v2 significantly outperform transformers by more than 10\%. Appendix \ref{sec:additional_icl_results} provides a similar comparison on a smaller model size ($\sim$1.5\textsc{b}), with an even larger margin. Appendix \ref{apx:model_efficiency} further summarizes the comparison under matched model size or computation (FLOPs). 

\begin{figure}[ht!]
    \centering
        \vspace{-1ex}
         \includegraphics[width=0.92\linewidth]{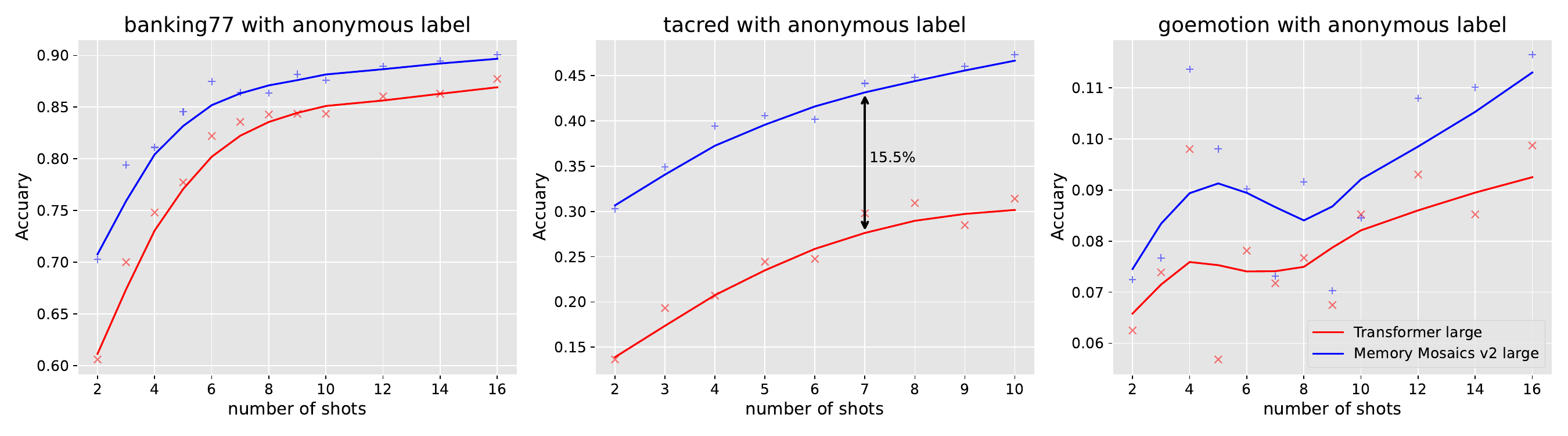}
             \vspace{-1.5ex}
    \caption{Anonymous label in-context learning comparison between memory mosaics v2 and transformers. Memory mosaics v2 significantly outperform transformers on all classification tasks.}
    \label{fig:inference-time-classification_anonymous_labels}
\end{figure}

Figure \ref{fig:inference-time-classification_anonymous_labels} presents a similar comparison as Figure \ref{fig:inference-time-classification_semantic_labels}, but on anonymous target labels. Again, memory mosaics v2 significantly outperforms transformers on all classification tasks.

In summary, the experiments demonstrate that memory mosaics v2 not only outperform transformer by a significant margin (more than 10\%) on in-context learning, but also consistently improve performance as more demonstrations are provided. These results highlight the superior in-context learning ability of Memory Mosaics v2.

\begin{wrapfigure}{r}{0.36\textwidth}
\vspace{-4ex}
  \begin{center}
    \includegraphics[width=0.32\textwidth]{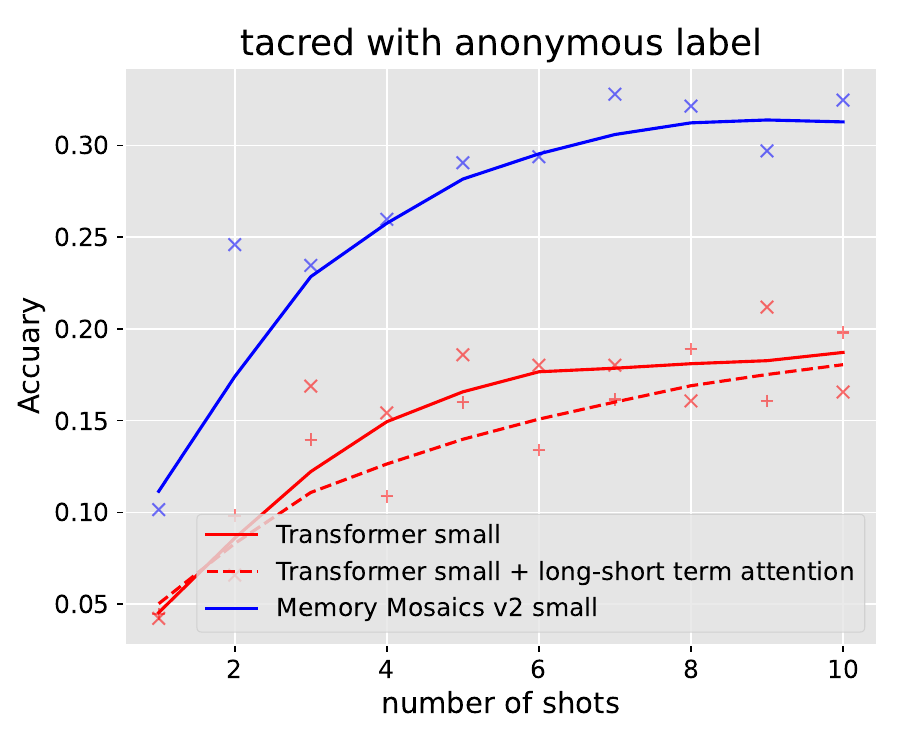}
  \end{center}
  \vspace{-3ex}
  \caption{Augmenting transformer with long-short term attention doesn't help in-context learning.}
  \label{fig:tf_ablation-anonymous}
  \vspace{-3ex}
\end{wrapfigure}

\paragraph{Augment transformer with long-short term attention} Memory mosaics (v2) contains several unique components that are not applicable to transformers, such as the symmetric key and query, and the adaptive bandwidth. One seemingly applicable component for transformers is the separation of long-term and short-term memories introduced in Section \ref{sec:3-level_memory}. However, Figure \ref{fig:tf_ablation-anonymous} shows that augmenting a transformer with long-short-term attention does not help it overcome the limitations of in-context learning. These phenomena imply that memory mosaics (v2) is not simply a transformer variation but represents a different architecture.

\paragraph{Computation and Parameter Concerns}
On the last two evaluation dimensions (new knowledge storage and retrieval, and in-context learning), memory mosaics v2 outperform transformers by more than 10\% with slightly more parameters. This 10\% advantage holds even when comparing under the same number of parameters or the same computational budget. See Appendix Figure \ref{fig:model_efficiency} for details.

\section{Risk-return trade-off of frontier-model-sized memory mosaics v2}
\label{sec:risk-return-trade-off-mmv2}
Having demonstrated the superior new tasks learning ability of memory mosaics v2 up to 9.9 billion parameters and 1 trillion training tokens, this section analyzes the ``risk-return trade-off'' to further scale memory mosaics v2 to the size of the frontier model, unveiling potential benefits and challenges.
\bfparagraph{Two Approaches} To train a large frontier foundational model, one can either: 
\begin{itemize}[leftmargin=.2in]
\vspace{-1ex}
     \setlength\itemsep{-0.3em}
    \item[1)] take a low-risk-low-return approach by investing more resources (GPUs and data) and reusing old recipes (e.g. architecture), or
    \item[2)] take a middle-risk-high-return approach by trying new smart techniques. 
\end{itemize}
Taking the first approach, one can take advantage of existing software, hardware, experiences, and datasets to quickly ``reproduce'' a huge foundational model. However, this approach is unlikely to result in a model that stands out from others, as it is based on shared recipes.

In contrast, taking the latter approach may require optimizing software and hardware, adapting techniques, a sharp sense of research direction, and possessing a keen sense of research direction along with strong problem solving abilities.\footnote{ {These requirements, in turn, demand a small group of high-quality researchers and managers.}} Despite the high requirements for personnel, this approach holds the potential for tremendous breakthroughs.

Ultimately, the decision between these two approaches depends on the available resources and personnel. To aid in this decision-making process, this section provides a simple and brutal comparison:
\begin{boxA}
\vspace{-1ex}  
    How much more data does the transformer recipe approach need to match the performance of memory mosaics v2?
    \vspace{-1ex}
\end{boxA}

\subsection{Comparison of two approaches}

To answer this question, we compare the new tasks learning ability\footnote{In \iid. regime, such as persistent-knowledge storing and retrieval, of course, more data + larger model = better performance. This argument in \iid.~scenario has been well studied three decades ago \cite{vapnik1991risk}.}  of memory mosaics v2 and transformers trained on various amounts of data. Specifically, multiple transformer models are trained on 200\textsc{b}, 1\textsc{t}, and 8\textsc{t} training tokens, while a memory mosaics v2 is trained on 1\textsc{t} training tokens.

\paragraph{{New-knowledge storage and retrieval}} Table \ref{tab:tf_mm_ruler_qa_tasks_size_of_data} shows the comparison on the {new-knowledge storage and retrieval} ability. Training on the same number of tokens (1\textsc{t}), transformers lag behind memory mosaics v2 by 12.3\% (41.1\% vs 53.4\%). $\times8$ times more training tokens (8\textsc{t}) improves the performance of transformers. However, the resulting transformer (trained on 8\textsc{t} tokens) still lags behind memory mosaics v2 (trained on 1\textsc{t} tokens) by 6.5\% (46.9\% vs 53.4\%).

Although further increasing training data may improve the performance of transformers in this evaluation dimension, it comes at the cost of significantly larger training cost (time and resource). Moreover, a serious problem occurs: we are running out of data!

\begin{table}[ht!]
    \vspace{-2ex}
    \centering
        \caption{Comparison of memory mosaics v2 and transformers, trained on 4k and fine-tuned on 32k context length, on \textsc{ruler} question-answer tasks. (``transformer large*'' uses group-query attention to reduce memory cost, increases training context length to 8k to boost long-context performance.)}
    \label{tab:tf_mm_ruler_qa_tasks_size_of_data}
    \vspace{0.5ex}
    \setlength{\tabcolsep}{2mm} 
    \resizebox{0.9\textwidth}{!}{
    \begin{tabular}{ccc|c c c c | c cc }
    \toprule
         model         &   \makecell{context length} & \makecell{train tokens}     &   \makecell{4k}   &   \makecell{8k}   &   \makecell{16k}   &   \makecell{task-length 32k} & \makecell{64k} && \\
         \midrule
transformer large    &32k & 200\textsc{b} &  48.6  &  42.9  &  40.7  &  33.8 & $\times$&&\\ 
transformer large    &32k & 1\textsc{t}   &  51.2  &  48.8  &  44.7  &  41.1 & $\times$&&\\
transformer large*    &32k & 8\textsc{t}  &  59.2  &  54.5  &  50.9  &  46.9 & $\times$&& \\
\midrule
memory mosaics v2 large &32k & 1\textsc{t} &  58.9  &  55.5  &  54.9  &  53.4 &46.4&\\
\bottomrule
    \end{tabular}
    }
\end{table}
\begin{figure}[ht!]
    \centering
    \includegraphics[width=0.92\linewidth]{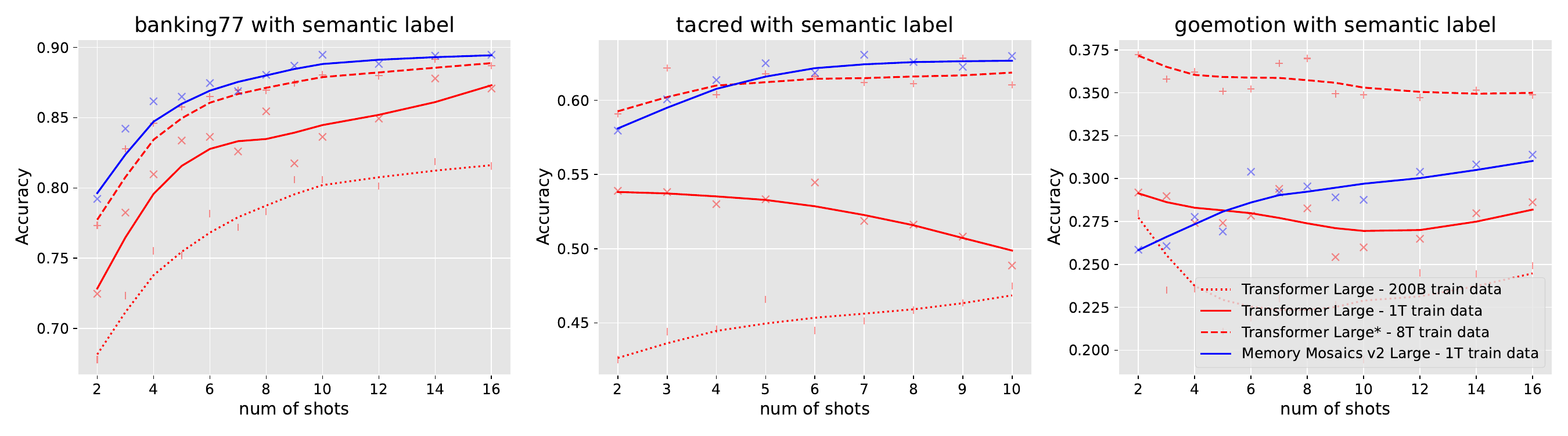}
    \vspace{-1ex}
    \caption{Semantic label in-context learning comparison between memory mosaics v2 and transformer.  Memory mosaics v2 is trained on 1\textsc{t} tokens, while three transformers are trained on 200\textsc{b}, 1\textsc{t}, 8\textsc{t} tokens, respectively. Transformer with $\times 8$ times more training data (8\textsc{t}, \textcolor{red}{dash red line}) starts to match the performance of Memory Mosaics v2 (1\textsc{t}, \textcolor{blue}{solid blue line}). }
    \label{fig:tf_mm_classification_semantic}
\end{figure}
\begin{figure}[ht!]
    \centering
    \includegraphics[width=0.92\linewidth]{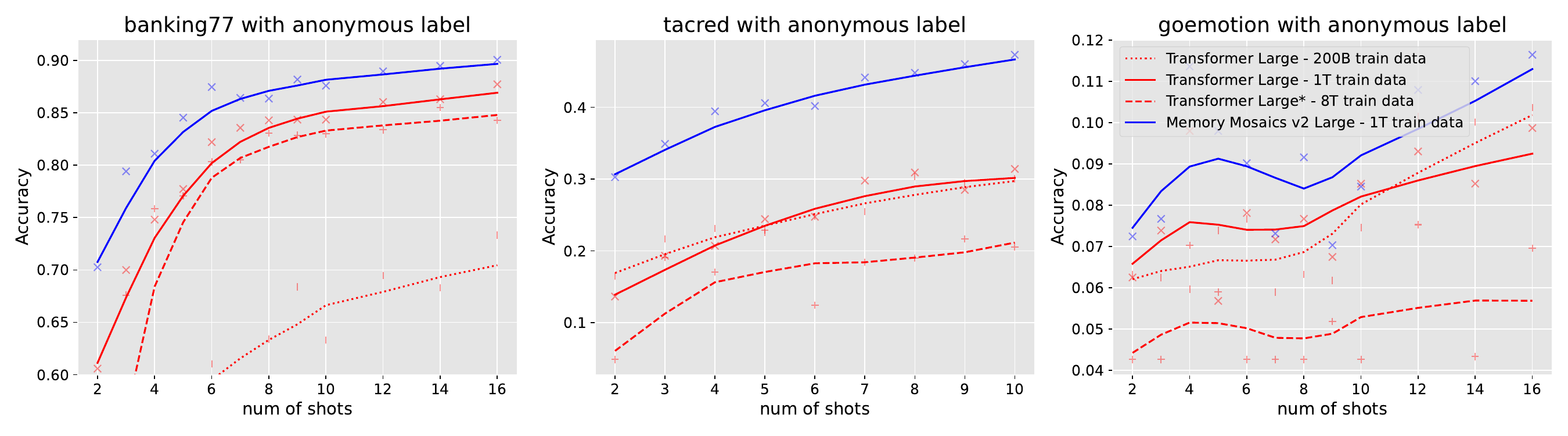}
    \vspace{-1ex}
    \caption{nonymous label in-context learning comparison between memory mosaics v2 and transformers. Sadly, transformers trained on 8\textsc{t} (\textcolor{red}{dash red line}) still lag behind memory mosaics v2 trained on 1\textsc{t} (\textcolor{blue}{solid blue line}) by a large margin. }
    \label{fig:tf_mm_classification_anonymous}
    \vspace{-2ex}
\end{figure}

\paragraph{In-context learning} Figures \ref{fig:tf_mm_classification_semantic} and \ref{fig:tf_mm_classification_anonymous} show the comparison on in-context learning ability. For semantic label tasks (Figure \ref{fig:tf_mm_classification_semantic}), $\times 8$ times more training data helps transformers (8\textsc{t} data) match the performance of memory mosaics v2 (1\textsc{t} data).  However, for the more challenging anonymous label tasks, more training data cannot help transformers. Contour-intuitively, transformers trained on more training data (8\textsc{t}) exhibit a degraded performance on anonymous label tasks (Figure \ref{fig:tf_mm_classification_anonymous}).

In summary, $\times 8$ more training data helps transformers in certain new task learning benchmarks. However, the resulting transformers (8\textsc{t} data) still lag behind memory mosaics v2 trained on 1\textsc{t} data. More importantly, in anonymous label tasks that heavily rely on the new task learning ability, more training data cannot help transformers. These experiments answer the initial question: \textit{``How much data does the transformer recipe approach need to match the performance of memory mosaics v2?''}.

\section{Fine-tuning speed: who can fine-tune with one minibatch?}
\label{sec:fine-tuning_speed}

Despite the strong in-context learning capability of memory mosaics v2 shown in Section \ref{sec:inference-data-learning}, it may still be attractive to fine-tune a model for a specific domain in order to either reduce inference costs or improve in-domain performance. It is generally expected that such models can be efficiently fine-tuned for a new domain using a comparatively small number of examples.

\begin{wrapfigure}{r}{0.4\textwidth}
\vspace{-2ex}
  \begin{center}
    \includegraphics[width=0.4\textwidth]{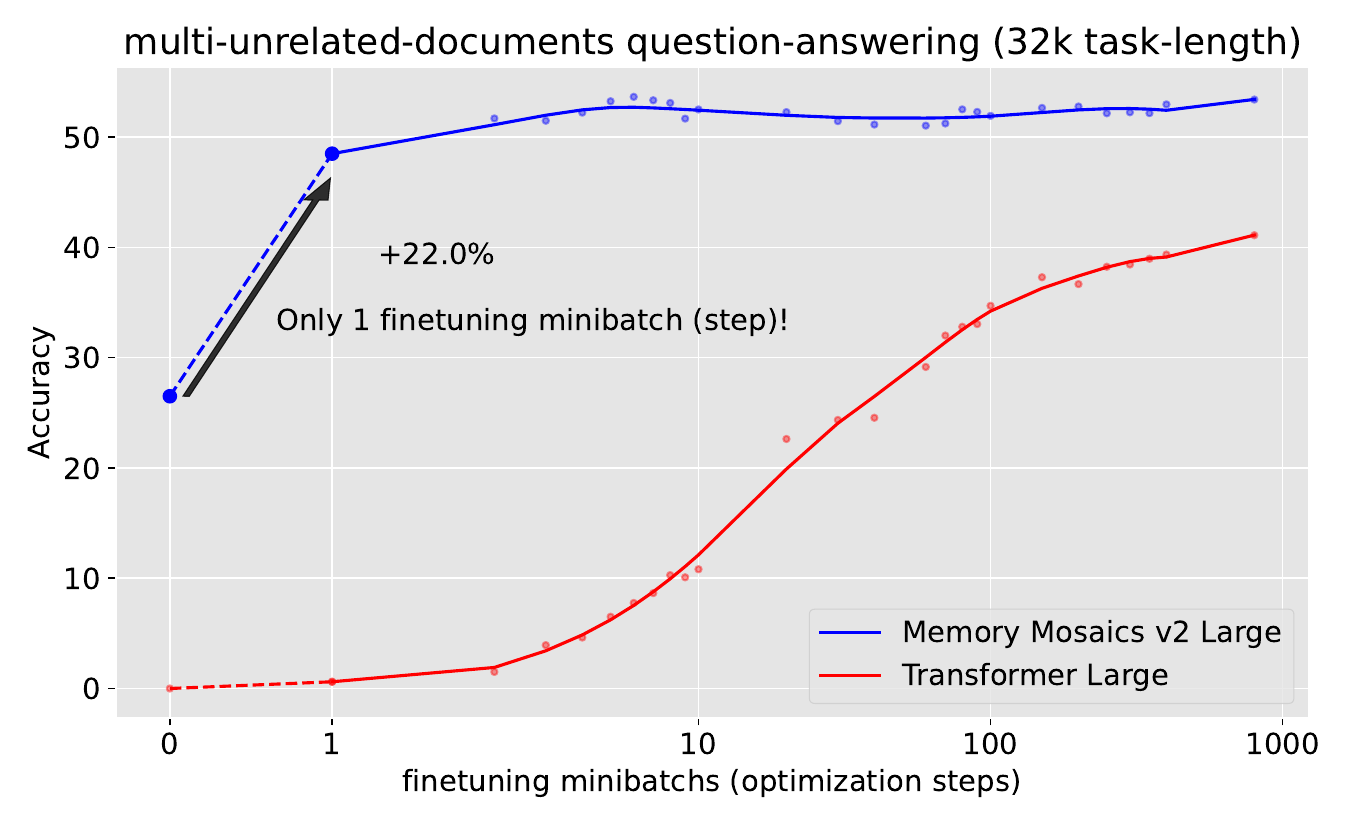}
  \end{center}
  \vspace{-2ex}
  \caption{Fine-tuning speed comparison between memory mosaics v2 and transformer.}
  \label{fig:finetune_speed}
  \vspace{-3ex}
\end{wrapfigure}

Figure \ref{fig:finetune_speed} compares the fine-tuning speed (in terms of data size) of memory mosaics v2 and transformers. Both models, pre-trained on 4k context windows, were fine-tuned to 32k context length using the recipe described in Section \ref{sec:mmv2_training} and evaluated on the same \textsc{ruler} tasks (32k task-length) described in Section \ref{sec:long-term-mem_utilization}.

Surprisingly, a single fine-tuning mini-batch (one optimization step) on memory mosaics v2 yields a 22\% accuracy improvement. Two fine-tuning mini-batches on memory mosaics v2 are sufficient to reach the optimal performance. In contrast, a transformer fine-tuned with 800 mini-batches still lags behind memory mosaics v2 fine-tuned with a single mini-batch.

\section{Discussion and future direction}
\label{sec:discussion}

This work scales memory mosaics (named memory mosaics v2) to llama-8\textsc{b} scale, demonstrating superior performance on new task learning, outperforming transformers by more than 10\%. The three evaluation dimensions introduced in this work provide a transparent and controlled assessment of model capabilities, particularly focusing on the new task learning. The risk-return trade-off analysis reveals the weakness of the mainstream ``more data more computation'' belief, highlighting research opportunities on other smart techniques. One future direction is to reduce the computational cost for very long context lengths using fuzzy hashing \cite{breitinger2014approximate, chen2024magicpig} and hierarchical memory \cite{yuan2025native, lu2025moba} approaches.

\appendix

\renewcommand{\acksection}{\section*{Acknowledgments}}
\begin{ack}
L\'eon Bottou is a CIFAR fellow. \Jupdate{We thank Gabriel Synnaeve, Jade Copet, Badr Youbi Idrissi, and Ammar Rizvi for their considerable support with hardware, software, data, and baselines.}

\end{ack}

\newpage
\bibliographystyle{plainnat}
\bibliography{neurips_2024}

\newpage
\hrule
\begin{center}
\LARGE Memory Mosaics at scale
\end{center}

\begin{center}
\large Supplementary Material
\end{center}

\hrule
\vskip 1cm

\section{Gated time-variant key feature extractor \& convolutional value extractor }

\begin{figure}[ht!]
    \centering
    \includegraphics[width=0.8\linewidth]{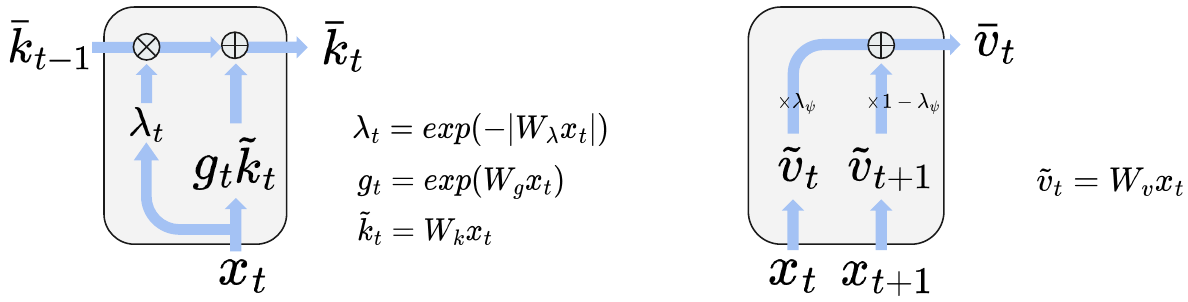}
    \caption{\textbf{Left:} Key $k_t$ feature extractor: $k_t = \text{Norm}(\bar{k}_t)$. \textbf{Right:} Value $v_t$ feature extractor: $v_t = \alpha_{\psi}\text{Norm}(\bar{v}_t)$.}
    \label{fig:key-feature-extractor-gated}
\end{figure}

Figure \ref{fig:key-feature-extractor-gated} illustrates how keys and values are constructed in memory mosaics v2. 

\section{Training data sequence length distributions}

\begin{figure}[h!]
    \centering
    \includegraphics[width=0.48\linewidth]{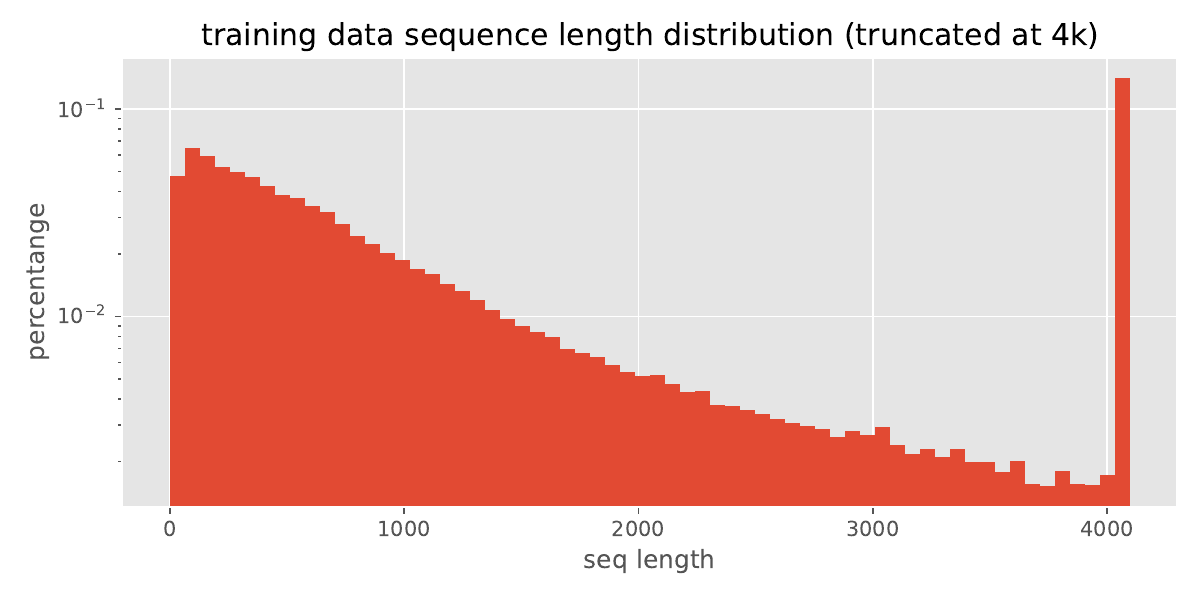}
    \includegraphics[width=0.48\linewidth]{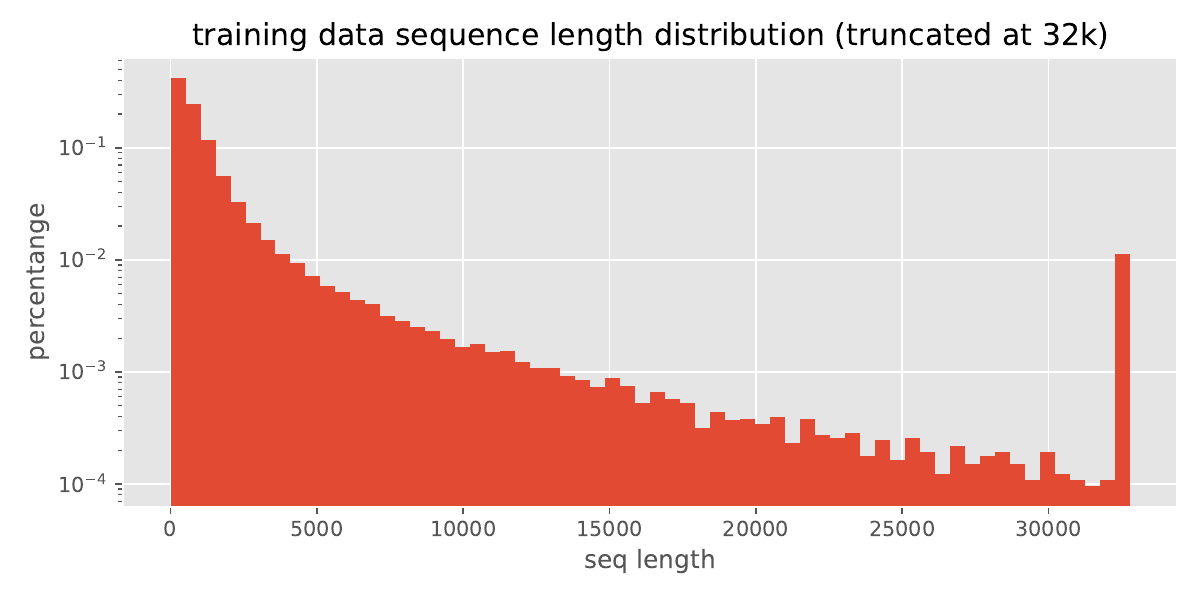}
    \caption{Training data sequence length distributions. For a given maximum sequence length during training (e.g. 4k), longer sequences are truncated to the maximum sequence length. This truncation results in the peaks at the end of distributions. }
    \label{fig:mm_training_data_seq_lens}
\end{figure}

Figure \ref{fig:mm_training_data_seq_lens} shows the distributions of the length of the training data sequence truncated to 4096 or 32,768 max length. 

\clearpage

\section{Training details}
\label{sec:training_details}
\bfparagraph{hyperparameters}

For all Memory Mosaics v2 and baseline Transformer models\footnote{
The hidden dimension of the two-layers neural network in persistent memory is set to 6,144 and 14,336 for small and large models, respectively.}, we use a consistent set of hyperparameters. That is, a batch size of 1024, a sequence length of 4096, an adamw optimizer with $\beta_1=0.9$  and $\beta_2=0.95$ accompanied by a $L_2$ weight decay of 0.1 and a gradient norm clip of 1, a learning rate warm-up of 2000 iterations followed by a cosine learning rate scheduler that reduces the learning rate by a factor of 100 at the end. The initial learning rates (after warm-up) are set to 3e-4 for ``small'' models and 1e-3 for ``large'' models.

We also employ document-wise attention mask, where the attention scores are only computed within each sequence (document) in the training data, to reduce computation cost. Two special tokens, ``<|begin\_of\_text|>'' and ``<|end\_of\_text|>'' are appended at the begining and ending of a sequence, respectively.

\begin{figure}[ht!]
    \centering
    \vspace{-1ex}
    \includegraphics[width=0.5\linewidth]{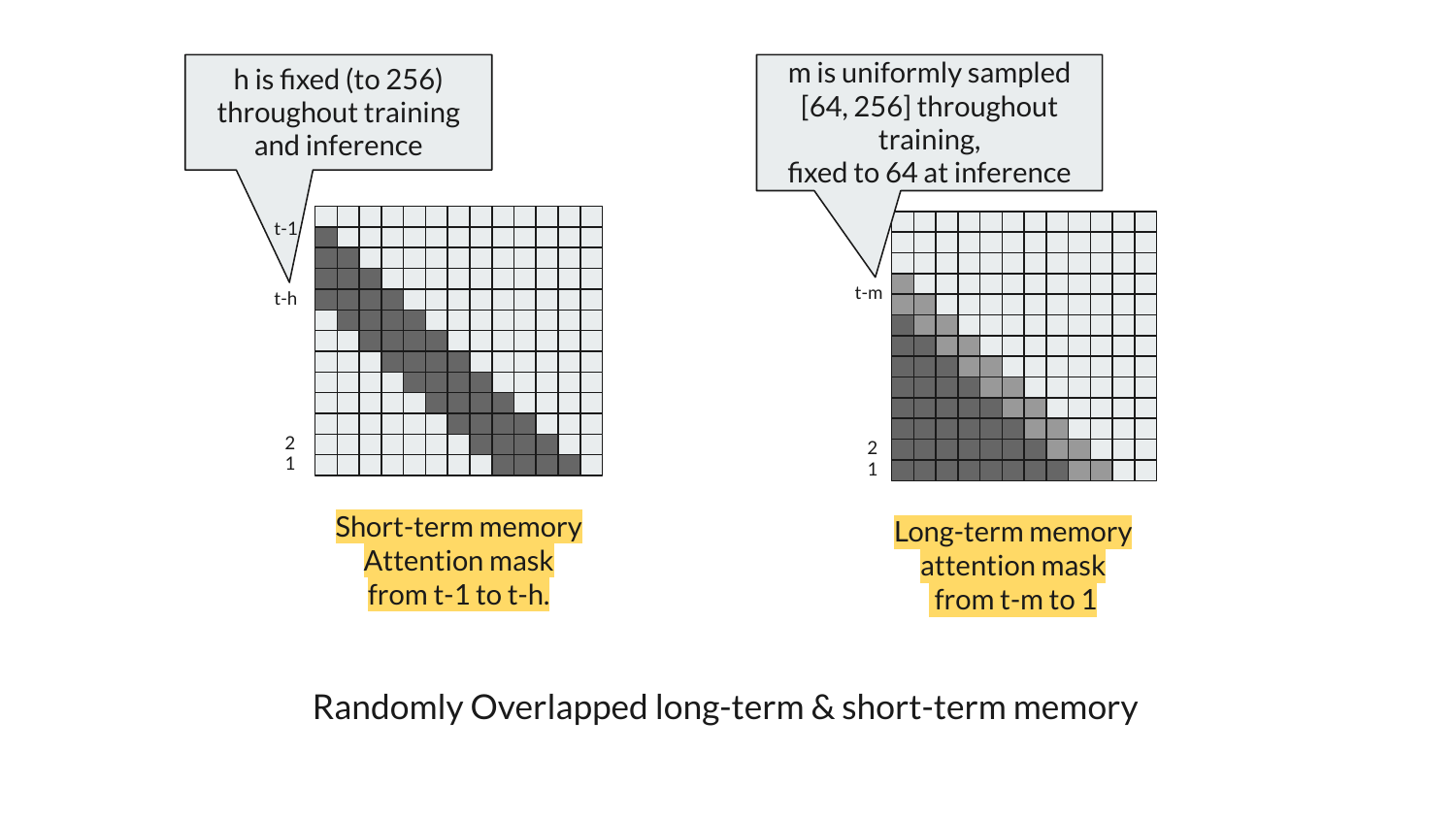}
    \vspace{-1ex}
    \caption{Randomly overlapped long-term \& short-term memory.}
    \label{fig:overlapped-long-short-memory}
    \vspace{-1ex}
\end{figure}

During training, memory mosaics v2 samples the long-term memory delay step $m$ from $[64, 256]$, sets the short-term memory window size $h=256$. At inference, $m$ is set to 64, as illustrated in Figure \ref{fig:overlapped-long-short-memory}. 

It is worth noting that these hyperparameters were originally searched and optimized for the baseline transformer models. We transfer these hyperparameters to memory mosaics v2 without further hyperparameter searching. Thus, it is possible that this hyperparameter setup is suboptimal for memory mosaics v2.

\Jupdate{
\paragraph{Parameter Initialization and reparameterization}
Table \ref{tab:param_init} summarizes the parameter initialization methods and reparameterization tricks. $W_1, W_2, W_3$ refer to the parameters in persistent memory that are implemented as two-layer dense neural networks, $W_2\big(SiLU(W_1(x)) * W3(x)\big)$.  $SiLU(x) = x \cdot sigmoid(x)$ is an activation function. $d \in \{2048, 4096\}$ indicates the hidden dimension of Memory Mosaics v2 small and large. $d^\prime \in \{6144, 14336\}$ indicates the hidden dimension of the two-layer neural networks in persistent memory. $l$ indicates the depth of the Mosaics blocks, starting from $0$. }
\begin{table}[ht!]
\setlength\extrarowheight{3pt}
\setlength{\tabcolsep}{10pt} 
    \centering
        \caption{Parameter initialization methods and reparameterization tricks used in Memory Mosaics v2.}
    \label{tab:param_init}
    \resizebox{1\textwidth}{!}{
    \begin{tabular}{c|ccc}
    \toprule
        Parameter & Location & reparameterization & Initialization  \\
         \midrule
         $\beta_0$ & adaptive bandwidth &  $\beta_0 = e^{\min(\theta, 10)}$ & $\theta=1.5$\\
         $\beta_1$ & adaptive bandwidth &  $\beta_1 = e^{\min(\theta, 10)}$ & $\theta=1.5$\\
         $\alpha$ & adaptive bandwidth & $\alpha = \min(|\theta|,1)$ &  $\theta=1/3$ \\
         $\alpha_{\psi}$ & feature extractor & $\alpha_{\psi} = e^{\min(|\theta|, 15)}$  & $\theta=0$ \\
         $\gamma$ & feature extractor & - & $U(0,1)$ \\
         $W_{\psi}, W_{\varphi}, W_{g}, W_{\lambda}, W_o$ &long-short memory & - & $\min\big(\max(\mathcal{N}(0, \sigma), -3\sigma), 3\sigma\big)$, $\sigma = \frac{1}{\sqrt{2d(l+1)}} $ \\
         $W_1, W_3$ & persistent memory & - & $\min\big(\max(\mathcal{N}(0, \sigma), -3\sigma), 3\sigma\big)$, $\sigma = \frac{1}{\sqrt{2d(l+1)}} $ \\
         $W_2$ & persistent memory & - & $\min\big(\max(\mathcal{N}(0, \sigma), -3\sigma), 3\sigma\big)$, $\sigma = \frac{1}{\sqrt{2d^\prime(l+1)}} $ \\
         $W_e, W_c$ & embedding \& classifier & - & $\min\big(\max(\mathcal{N}(0, \sigma), -3\sigma), 3\sigma\big)$, $\sigma = \frac{1}{\sqrt{2d}} $\\
         \bottomrule
    \end{tabular}}
\end{table}

\clearpage
\section{Failures of memory compression baselines}
\label{apx:fail_mem_compression}

\Jupdate{Many memory compression algorithms, such as \textsc{rnn}s, x\textsc{lstm} \citep{beck2025xlstm}, rwkv \citep{peng2023rwkv}, and state-space models \citep{gu2023mamba}, fail on \textbf{new-task storage and retrieval} and \textbf{in-context learning} evaluation dimensions by construction. The reason is that these memory compression algorithms lack the ability to store large amounts of information before getting a command on how to process the information. One might argue to play around this shortage by reading the ``command'' before storing the large amounts of information. However, this process involves task-specific priori knowledge from human designers. In the end, instead of proving the machine is intelligent, it often proves that human designers are intelligent. Please recall that \emph{a child does not prepare all questions before going to school}.}

\Jupdate{This incompetent of memory compression algorithms has been experimentally demonstrated by \citet{hsieh2024ruler} and \citet{li2024long} on both \textsc{ruler} benchmarks and in-context learning tasks.}

\Jupdate{Table \ref{tab:memory_compression_baselines_on_ruler} compares memory compression methods (rwkv-v5-7b and mamba-2.8b-slimpj) and non-compression method (llama2-7b) on \textsc{ruler} long-context tasks. It is clear that memory compression methods perform poorly as the required context length (i.e., required information storage space) increases.  }

\Jupdate{Similarly, Table \ref{tab:memory_compression_baselines_on_tacred} compares memory compression methods (rwkv-5-world 7b and Mamba-2.8B) and non-compression method (qwen-1.5-7b-base and mistral-7b-v0.2-base) on in-context learning tasks (Tacred few-shot classification \citep{zhang2017tacred}). In this challenging in-context scenario, memory compression methods just don't work at all.}

\Jupdate{Please note that this section shouldn't be used to criticize or hinder the study of memory compression methods. Memory compression methods have their advantages. In \textbf{persistent-knowledge storage and retrieval} evaluation dimension, they performs very well. For model efficiency, memory compression methods reveal a charming computation complexity. The goal of this section is to explain why this paper doesn't choose memory compression methods as baselines. }

\begin{table}[ht]
    \centering
     \caption{Comparison of memory compression methods (rwkv-v5-7b and mamba-2.8b-slimpj) and non-compression method (llama2-7b) on \textsc{ruler} long-context tasks. memory compression methods perform poorly as the required context length increases. Numbers are copied from \cite{hsieh2024ruler} Figure 4. }
\label{tab:memory_compression_baselines_on_ruler}
    \resizebox{0.6\textwidth}{!}{
    \begin{tabular}{c|ccc}
    \toprule
model	& task-length 1k &	task-length 2k &	task-length 4k \\
\midrule
llama2-7b&	96.0&	91.6&	95.0 \\
rwkv-v5-7b&	87.5&	73.7&	51.4 \\
mamba-2.8b-slimpj&	62.6&	52.6	&- \\
\bottomrule
    \end{tabular}}
\end{table}

\begin{table}[ht]
    \centering
    \caption{Comparison of memory compression methods (rwkv-5-world 7b and Mamba-2.8B) and non-compression method (qwen-1.5-7b-base and mistral-7b-v0.2-base) on in-context learning tasks (Tacred few-shot classification \citep{zhang2017tacred}). Memory compression methods fail on all cases. Numbers are copied from \cite{li2024long} Table 4. }
\label{tab:memory_compression_baselines_on_tacred}
    \resizebox{0.6\textwidth}{!}{
    \begin{tabular}{c|ccccc}
    \toprule
         model &	1-shot & 2-shots & 3-shots & 4-shots &	 5-shots \\
         \midrule
qwen-1.5-7b-base 7b &	38.7&	47.3&	45.2&	43.6&	40.6 \\
mistral-7b-v0.2-base&	53.3&	53.1&	51.6&	48.0&	42.3 \\
rwkv-5-world 7b	    &   2.3&	2.6&	1.0&	0&	1.2 \\
Mamba-2.8B          &	0&	0&	0&	0&	0 \\
    \bottomrule
    \end{tabular}}
\end{table}

\clearpage
\section{Model efficiency comparison}
\label{apx:model_efficiency}
\begin{figure}[ht]
    \centering
    \includegraphics[width=0.8\linewidth]{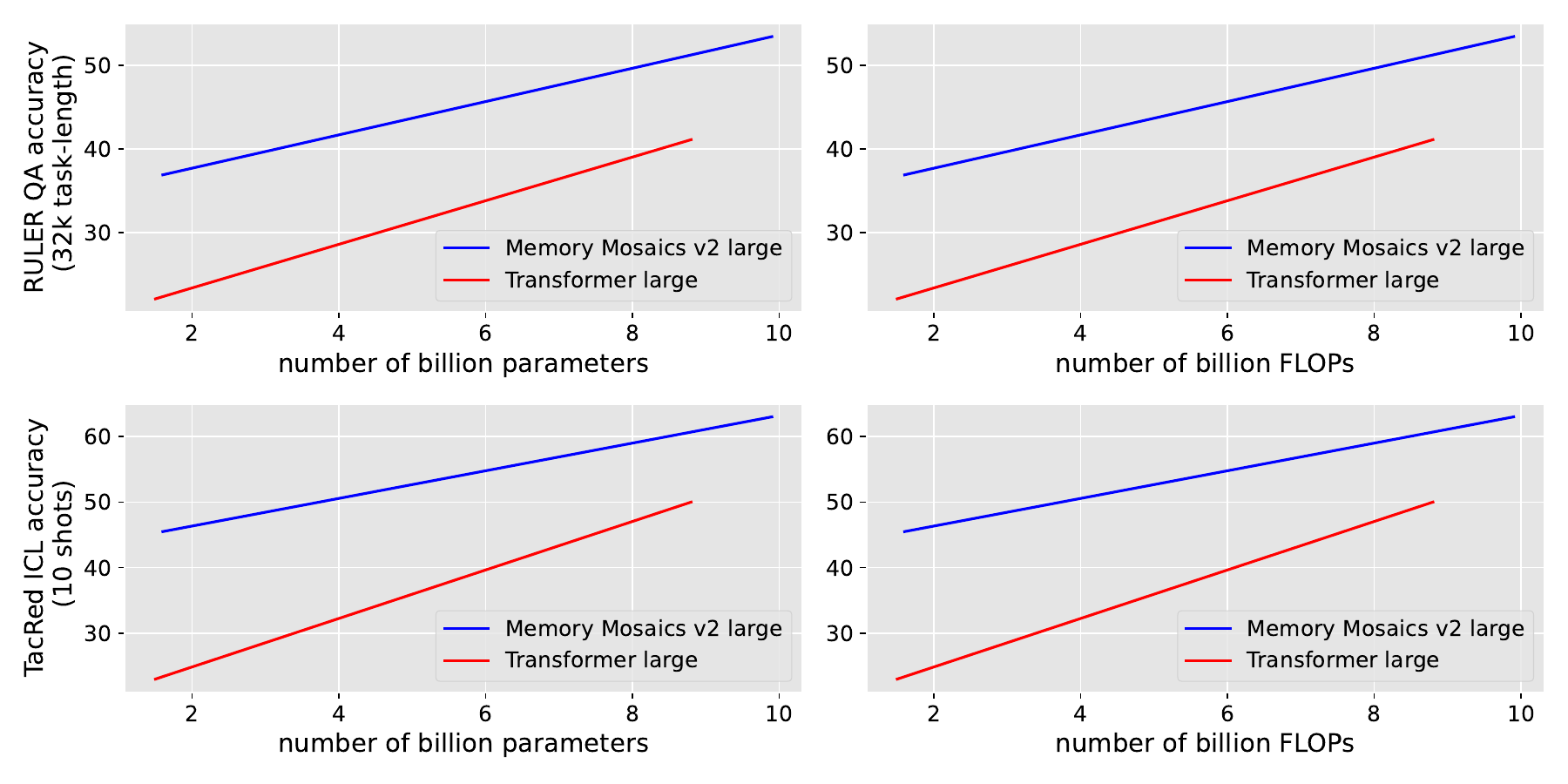}
    \caption{Model efficiency (FLOPs or \# parameters) comparison between Transformer large and Memory Mosaics v2 large. Top row shows the comparison on \textsc{ruler} ``multi-unrelated-documents storing and question-answering'' tasks, while the bottom row shows comparison on the Tacred in-context learning task. Memory mosaics v2 large outperforms transformer large by more than 10\%. }
    \label{fig:model_efficiency}
\end{figure}

\Jupdate{%
As we emphasized in main text, model efficiency (e.g. model service, throughput, \textsc{vram}, etc.) is not the goal of this work. Many engineering works can be performed to adapt memory mosaics v2 to a custom use case or hardware. To aid in these potential adaptations, Figure \ref{fig:model_efficiency} provides a model efficiency comparison in both computation (FLOPs) and model size (number of parameters) viewpoints. The results show that memory mosaics v2 outperforms transformer by more than 10\% under either the same FLOPs or the parameters budgets. 
}

\section{Additional results on persistent-knowledge storage and retrieval}

Table \ref{tab:6_normal_benchmarks_with_without_longterm_mem} shows six language benchmarks in which removing long-term memory from memory mosaics v2 after training degrades its performance.

\begin{table}[ht!]
    \centering
       \caption{Memory mosaics v2 performance on 6 language benchmarks, where removing the ``long-term memory'' after training dramatically hurt the performance (42.1\% vs 34.9\%).}  
    \label{tab:6_normal_benchmarks_with_without_longterm_mem}
    \resizebox{0.7\textwidth}{!}{
    \begin{tabular}{c|cc|ccc ccc  | c  }
    \toprule
  &  params & flops/token  &    squad & \makecell{bbh}  &  \makecell{math}  &  \makecell{mbpp}   &  \makecell{race\\ middle}  & \makecell{race \\ high} &  avg  \\
         \midrule
         transformer large  & 8.8B & 16.7B  &  76.3  &  45.6  &  8.7  &  9.8  &  62.6  &  45.6  &  41.4 \\
         \midrule
          \makecell{memory mosaics v2 large}  & 9.9B &  18.9B & 78.2  &  47.8  &  8.8  &  9.6  &  61.6  &  46.5  &  42.1 \\
        \makecell{memory mosaics v2 large \\without long-term memory} & 8.3B & 15.6B  &  69.4  &  24.6  &  5.4  &  6.8  &  59.5  &  43.6  &  34.9 \\
         \bottomrule
    \end{tabular}
    }

\end{table}

\clearpage
\section{Additional results on new-knowledge storage and retrieval}
\label{apx:new-knowledge}
Table \ref{tab:tf_mm_32k_ruler_qa_tasks_long-term-mem} shows that removing long-term memory from memory mosaics v2 after training degrades the performance on the \textsc{ruler} question-answer tasks by 20\%$\sim$30\%. This indicates that the ruler question-answer tasks rely on long-term memory to perform well. 

\Jupdate{Table \ref{tab:memory_mosaics_v2_and_other_base_models_on_ruler} compares memory mosaics v2 large and other public base models on \textsc{ruler} question-answer tasks. Memory mosaics v2 large outperforms these models across all task lengths.}

\Jupdate{Table \ref{tab:stochastic_long_mem} illustrates the effect of the stochastic long-term memory size training setup introduced in Section \ref{sec:training_details}. This stochastic long-term memory size setup is used to encourage the allocation of position-invariant signals and position-dependent signals to long-term and short-term memories.} 

\Jupdate{Table \ref{tab:ruler_s_niah} compares memory mosaics v2 and transformers on a typical `needle-in-a-haystack' task from \textsc{ruler} \citep{hsieh2024ruler}. The typical `needle-in-a-haystack' is too easy such that many models can achieve a near-perfect performance.}
\begin{table}[h!]
    \centering
    \vspace{-1ex}
    \caption{The effect of removing ``long-term memory'' of memory mosaics V2 large on \textsc{ruler} question-answer tasks.}
    \label{tab:tf_mm_32k_ruler_qa_tasks_long-term-mem}
    \resizebox{0.9\textwidth}{!}{
    \setlength{\tabcolsep}{4mm} 
    \begin{tabular}{cc|c c c c  }
    \toprule
         model          &   \makecell{context length}      &   4k   &   8k   &   16k   &   32k  \\
         \midrule
memory mosaics large  &32k & 58.9  &  55.5  &  54.9  &  53.4 \\
memory mosaics large without long-term memory  &32k &   38.5  &  22.2  &  20.0  &  20.2 \\
\bottomrule
    \end{tabular}
    }
    \vspace{-1ex}
\end{table}

\begin{table}[h!]
    \centering
    \vspace{-1ex}
    \caption{Comparison of Memory Mosaics v2 large (base model) and other public base models (similar scale) on \textsc{ruler} question-answer tasks. Memory Mosaics v2 large outperforms these models across all task lengths, despite that Memory Mosaics v2 uses 1/4  generation lengths (32 tokens) of other public base models (128 tokens). The numbers in ``*'' rows come from \citet{hsieh2024ruler}.   }
    \label{tab:memory_mosaics_v2_and_other_base_models_on_ruler}
    \setlength{\tabcolsep}{2mm} 
    \resizebox{0.9\textwidth}{!}{
    \begin{tabular}{@{}l|c|cccc}
    \toprule
    Model & claimed length & task-length 4k &  8k &  16k & 32k \\
    \midrule
    Memory-Mosaics-v2-large (base) &32k &  58.9  &  55.5  &  54.9  &  53.4 \\
    \midrule
    Llama2-7B (base)* & 4k & 48.6 & - & - & -  \\
Mixtral-base (8x7B)* & 32k & {50.8} & 47.7 & 45.3 & 41.3 \\
Mistral-base (7B)* & 32k & {53.5} & {51.0} & 48.4 & 44.7 \\
Together-base (7B)* & 32k  & 47.5 & 44.6 & 33.6 & 0.0  \\
LongLoRA-base (7B)* & 100k &  34.5 & 32.1 & 33.6 & 29.4 \\
Yarn-base (7B)* & 128k &  29.7 & 23.5 & 28.6 & 29.7  \\
LWM-base (7B)* & 1M & 42.7 & 40.2 & 38.7 & 37.1  \\
\bottomrule
    \end{tabular}}
\end{table}

\begin{table}[h!]
    \centering
     \vspace{-1ex}
        \caption{The effect of stochastic long-term memory size (during training) in memory mosaics v2 small model on \textsc{ruler} question-answer tasks. Both models are trained on 4k context length, then evaluated on 32k context length without any fine-tuning. The stochastic long-term memory size setup boost context length extrapolation ability by more than 15\%.}
    \label{tab:stochastic_long_mem}
    \resizebox{0.98\textwidth}{!}{
    \setlength{\tabcolsep}{2mm} 
    \begin{tabular}{ccc|cc  }
    \toprule
         model         &   \makecell{context length}  & stochastic long-term memory    &   \makecell{task-length 4k}   &      \makecell{task-length  32k}  \\
         \midrule
       memory mosaics v2 small & 4k &   No   & 43.6  &    15.9 \\
       memory mosaics v2 small & 4k &   Yes  & 45.0  &    \textbf{31.7 (+15.8)} \\
       
\bottomrule
    \end{tabular}
    }
     \vspace{-1ex}
\end{table}

\begin{table}[h!]
    \centering
    \caption{\textsc{ruler} \textsc{s-niah} benchmark comparison between transformer and memory mosaics v2.}
    \label{tab:ruler_s_niah}
    \resizebox{0.8\textwidth}{!}{
    \begin{tabular}{cc|ccc|c}
    \toprule
         model	& context length & 4k	& 8k & 16k 	& task-length 32k \\
         \midrule
transformer small & 32k &	99.4	& 99.0	& 98.2	& 97.8 \\
memory mosaics v2 small	& 32k & 100.0 &	100.0 &	100.0 &	100.0   \\
\midrule
transformer large & 32k &	100.0 &	100.0	 &100.0 &	99.6 \\
memory mosaics v2 large & 32k &	100.0 &	100.0 &	100.0	&100.0 \\
\bottomrule
    \end{tabular}}
\end{table}

\clearpage

\section{Additional results for in-context learning}
\label{sec:additional_icl_results}
\begin{figure}[h!]
    \centering
        \vspace{-1ex}
    \includegraphics[width=0.93\linewidth]{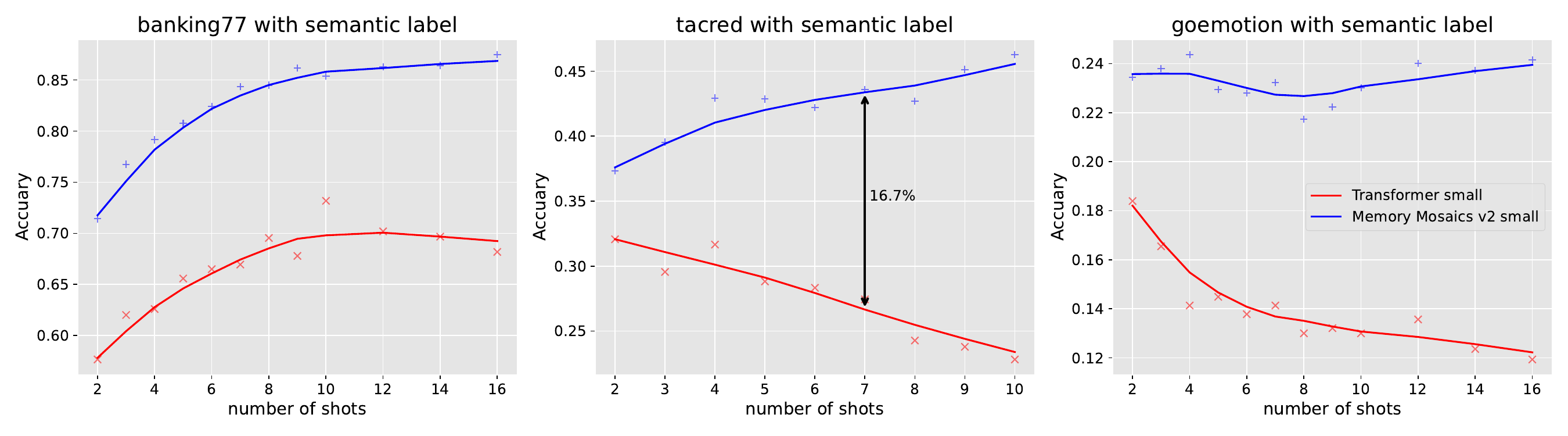}
        \vspace{-1ex}
    \caption{Semantic label in-context learning comparison between memory mosaics v2 and transformer. Memory mosaics v2 significantly outperform transformers on in-context learning with a large margin (more than 10\%). Meanwhile, memory mosaics v2 benefits from more demonstration shots (x-axis), unlike transformers.}
    \label{fig:inference-time-classification_semantic_labels_small}
        \vspace{-1ex}
\end{figure}

\begin{figure}[h!]
    \centering
        \vspace{-1ex}
    \includegraphics[width=0.93\linewidth]{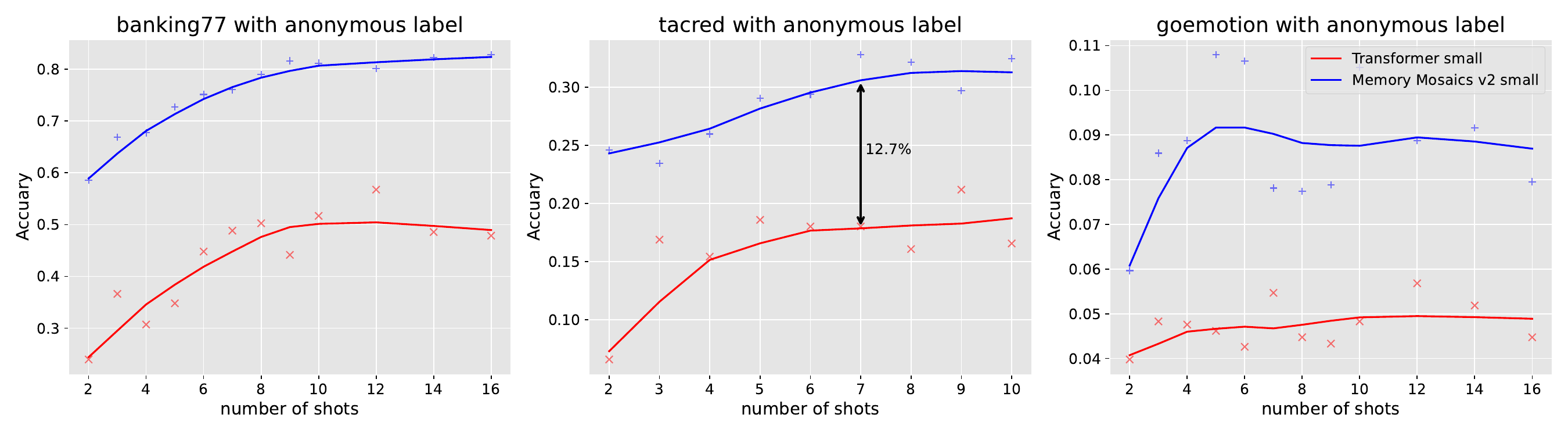}
             \vspace{-1ex}
    \caption{Anonymous label in-context learning comparison between memory mosaics v2 and transformers. Memory mosaics v2 significantly outperform transformers on all classification tasks.}
    \label{fig:inference-time-classification_anonymous_labels_small}
    \vspace{-1ex}
\end{figure}

Figure \ref{fig:inference-time-classification_semantic_labels_small} and \ref{fig:inference-time-classification_anonymous_labels_small} shows the in-context learning comparison between memory mosaics v2 small and transformer small (llama-1.5\textsc{b} scale).

\clearpage

\section{Prompt examples of multiclass classification tasks}
\label{apx:prompt_examples_icl_class}

\subsection{Banking77 classification with semantic labels}
We sweep the delimiter from  ``[return]'' and ``[space]'', leads to the following two prompts:

\begin{boxA}
``Given a customer service query, please predict the intent of the query. The predict answer must come from the demonstration examples with the exact format. The examples are as follows: 
\bigbreak
service query: \\
I am still waiting on my card? \\
intent category: \\
city\_arrival
\bigbreak
service query:\\
My card has been found. Is there any way for me to put it back into the app?\\
intent category:\\
city\_linking 
\bigbreak
...
\bigbreak
service query:\\
Can I get a card even if I live outside the UK?\\
intent category:\\''
\end{boxA}

\begin{boxA}
``Given a customer service query, please predict the intent of the query. The predict answer must come from the demonstration examples with the exact format. The examples are as follows: \\
service query: I am still waiting on my card? \\
intent category: city\_arrival \\
service query: My card has been found. Is there any way for me to put it back into the app?\\
intent category:
city\_linking \\ 
... \\
service query: Can I get a card even if I live outside the UK? \\
intent category:''
\end{boxA}

For each prompt with either ``[return]'' or ``[space]'' delimiter, we also try to shuffle the demonstration example (i.e., \textit{service query: [...], intent category:[...]}) orders within each one shot. This shuffling process provides another two more prompts. 

\subsection{Banking77 classification with anonymous labels}
Anonymous tasks use the same set of prompts except that anonymous tasks replace semantic labels (e.g. \textit{city\_arrival, city\_linking}) with anonymous labels (e.g. \textit{class\_00, class\_01}).

\clearpage

\subsection{Goemotion classification with semantic labels}
We sweep the delimiter from  ``[return]'' and ``[space]'', leads to the following two prompts: 

\begin{boxA}
``Given a comment, please predict the emotion category of this comment. The predict answer must come from the demonstration examples with the exact format. The examples are as follows: 
\bigbreak
comment:\\
Her upper lip always looks terrible - such an easy fix, can u believe she is so vain and never bothers to wax \\
emotion category:\\
embarrassment
\bigbreak
comment:\\
No problem. I'm happy to know it's not what you meant.\\
emotion category:\\joy
\bigbreak
...
\bigbreak
comment:\\
These refs have it out for the colts. I didn't realize we traded our MVP 11 to
KC either.\\
emotion category:\\''
\end{boxA}

\begin{boxA}
``Given a comment, please predict the emotion category of this comment. The predict answer must come from the demonstration examples with the exact format. The examples are as follows: \\
comment:
Her upper lip always looks terrible - such an easy fix, can u believe she is so vain and never bothers to wax \\
emotion category:
embarrassment \\
comment: No problem. I'm happy to know it's not what you meant.\\
emotion category: joy
...
comment: These refs have it out for the colts. I didn't realize we traded our MVP 11 to
KC either.\\
emotion category:''
\end{boxA}
For each prompt with either ``[return]'' or ``[space]'' delimiter, we also try to shuffle the demonstration example orders within each one shot. This shuffling process provides another two more prompts. 

\subsection{Goemotion classification with anonymous labels}
Anonymous tasks use the same set of prompts except that anonymous tasks replace semantic labels with anonymous labels (e.g. \textit{class\_00, class\_01}).

\clearpage
\subsection{Tacred classification with semantic labels}
We sweep the delimiter from  ``[return]'' and ``[space]'', leads to the following two prompts: 

\begin{boxA}
``Given a sentence and a pair of subject and object entities within the sentence, please predict the relation between the given entities. The examples are as follows: 
\bigbreak
sentence:\\
But US and Indian experts say it has hesitated to take action against Lashkar-e-Taiba, which means ``The Army of the Pure, ''believing that the Islamic militants could prove useful in pressuring its historic rival India.\\
the relation between Lashkar-e-Taiba and Army of the Pure is:\\
org:alternate\_names
\bigbreak
sentence:\\
The offer from ITW, the Glenview, Ill, diversified manufacturer of engineered products, represents a premium of 85 percent to the Manitowoc bid.
\\the relation between ITW and Glenview is:\\
org:city\_of\_headquarters
\bigbreak
...
\bigbreak
sentence:\\
The statement from North Korea, carried by the country's official Korean Central News Agency, did not mention Kim by name, but South Korean Unification Ministry spokesman Kim Ho-nyeon said the North's state media has before used such wording to refer to him.\\
the relation between Korean Central News Agency and North Korea is:\\''
\end{boxA}

\begin{boxA}
``Given a sentence and a pair of subject and object entities within the sentence, please predict the relation between the given entities. The examples are as follows: \\
sentence:
But US and Indian experts say it has hesitated to take action against Lashkar-e-Taiba, which means ``The Army of the Pure, ''believing that the Islamic militants could prove useful in pressuring its historic rival India.\\
the relation between Lashkar-e-Taiba and Army of the Pure is:
org:alternate\_names
\\
sentence:
The offer from ITW, the Glenview, Ill, diversified manufacturer of engineered products, represents a premium of 85 percent to the Manitowoc bid.
\\
the relation between ITW and Glenview is:
org:city\_of\_headquarters
\\
...
\\
sentence:
The statement from North Korea, carried by the country's official Korean Central News Agency, did not mention Kim by name, but South Korean Unification Ministry spokesman Kim Ho-nyeon said the North's state media has before used such wording to refer to him.\\
the relation between Korean Central News Agency and North Korea is:''
\end{boxA}

For each prompt with either ``[return]'' or ``[space]'' delimiter, we also try to shuffle the demonstration example orders within each one shot. This shuffling process provides another two more prompts. 

\subsection{Tacred classification with anonymous labels}
Anonymous tasks use the same set of prompts except that anonymous tasks replace semantic labels with anonymous labels (e.g. \textit{class\_00, class\_01}).

\section{Computation resources}
\label{sec:computation_resources}
All experiments are conducted on H100 GPUs with 80GB VRAM.

\end{document}